\documentclass{article}

\usepackage[preprint,nonatbib]{neurips_2023}

\usepackage[utf8]{inputenc} 
\usepackage[T1]{fontenc}    
\usepackage{url}            
\usepackage{booktabs}       
\usepackage{amsfonts}       
\usepackage{nicefrac}       
\usepackage{microtype}      

\usepackage{microtype}
\usepackage{graphicx}
\usepackage{booktabs} 
\usepackage{tabularx}

\usepackage{amsmath}
\usepackage{amssymb}
\usepackage{url}
\usepackage{multirow}
\usepackage{tablefootnote}
\usepackage{xspace}
\usepackage{xcolor,colortbl}
\usepackage{wrapfig}
\usepackage{makecell}

\usepackage{caption}
\usepackage{subcaption}

\definecolor{defaultcolor}{gray}{0.9}
\definecolor{citecolor}{HTML}{000080}
\definecolor{linkcolor}{HTML}{0000FF}
\usepackage[pagebackref=false, breaklinks=true, colorlinks, urlcolor=gray,citecolor=citecolor, linkcolor=linkcolor, bookmarks=false]{hyperref}

\usepackage[capitalize,noabbrev]{cleveref}
\crefname{section}{Sec.}{Secs.}
\Crefname{section}{Section}{Sections}
\Crefname{table}{Table}{Tables}
\crefname{table}{Tab.}{Tabs.}

\newlength\savewidth\newcommand\shline{\noalign{\global\savewidth\arrayrulewidth
\global\arrayrulewidth 1pt}\hline\noalign{\global\arrayrulewidth\savewidth}}
\newcommand{\tablestyle}[2]{\setlength{\tabcolsep}{#1}\renewcommand{\arraystretch}{#2}\centering\small}
\makeatletter\renewcommand\paragraph{\@startsection{paragraph}{4}{\z@}
{.4em \@plus1ex \@minus.2ex}{-.5em}{\normalfont\normalsize\bfseries}}\makeatother

\newcolumntype{x}[1]{>{\centering\arraybackslash}p{#1pt}}
\newcolumntype{y}[1]{>{\raggedright\arraybackslash}p{#1pt}}
\newcolumntype{z}[1]{>{\raggedleft\arraybackslash}p{#1pt}}
\newcolumntype{C}[1]{>{\centering\arraybackslash}p{#1}}

\makeatletter
\DeclareRobustCommand\onedot{\futurelet\@let@token\@onedot}
\def\@onedot{\ifx\@let@token.\else.\null\fi\xspace}
\def\eg{\emph{e.g}\onedot} 
\def\ie{\emph{i.e}\onedot} 
 
\def\etc{\emph{etc}\onedot}

\makeatother

\def\ourmodel{{MAViL}\xspace}

\definecolor{mygreen}{rgb}{0.0, 0.5, 0.0}

\title{MAViL: Masked Audio-Video Learners}

\author{%
Po-Yao Huang \quad Vasu Sharma \quad Hu Xu \quad Chaitanya Ryali \quad Haoqi Fan \quad Yanghao Li \quad\\
\textbf{Shang-Wen Li} \quad \textbf{Gargi Ghosh} \quad \textbf{Jitendra Malik} \quad \textbf{Christoph Feichtenhofer}
\\
FAIR, Meta AI
}

\begin{document}

\maketitle
\begin{abstract}
We present Masked Audio-Video Learners (\ourmodel) to learn audio-visual representations with three complementary forms of self-supervision: (1) reconstructing masked raw audio and video inputs, (2) intra-modal and inter-modal contrastive learning with masking, and (3) self-training to predict aligned and contextualized audio-video representations learned from the first two objectives.
Empirically, \ourmodel achieves state-of-the-art audio-video classification performance on AudioSet (53.3 mAP) and VGGSound (67.1\% accuracy), surpassing recent self-supervised models and supervised models that utilize external labeled data. 
Notably, pre-training with \ourmodel not only enhances performance in multimodal classification and retrieval tasks, but it also improves the representations of each modality in isolation, without relying on information from the other modality during uni-modal fine-tuning or inference.
The code and models will be available at \url{https://github.com/facebookresearch/MAViL}.
\end{abstract}

\section{Introduction}
\label{sec:intro}
We study self-supervised learning (SSL) from audio and video, two rich, heterogeneous, yet closely related modalities of human perception. 
There are two primary forms of self-supervision commonly used: reconstruction and contrastive learning.
By reconstructing masked text tokens on large-scale corpora, BERT~\cite{bert} pre-training has achieved groundbreaking results in various NLP tasks.
Masked autoencoders (MAEs) have recently emerged as powerful tools for learning uni-modal representations in various modalities such as image~\cite{mae}, video~\cite{feichtenhofer2022masked}, and audio~\cite{huang2022masked}. 
They employ an asymmetric encoder-decoder architecture with a substantial portion of encoder inputs being masked, resulting in efficient representation learning.
Additionally, contrastive learning has been widely used to achieve cross-modal alignment for image-text~\cite{clip,wei2022contrastive} and audio-visual~\cite{owens2018audio,stica} tasks.

\begin{figure}[t!]
    \centering
    \includegraphics[width=1.0\linewidth]{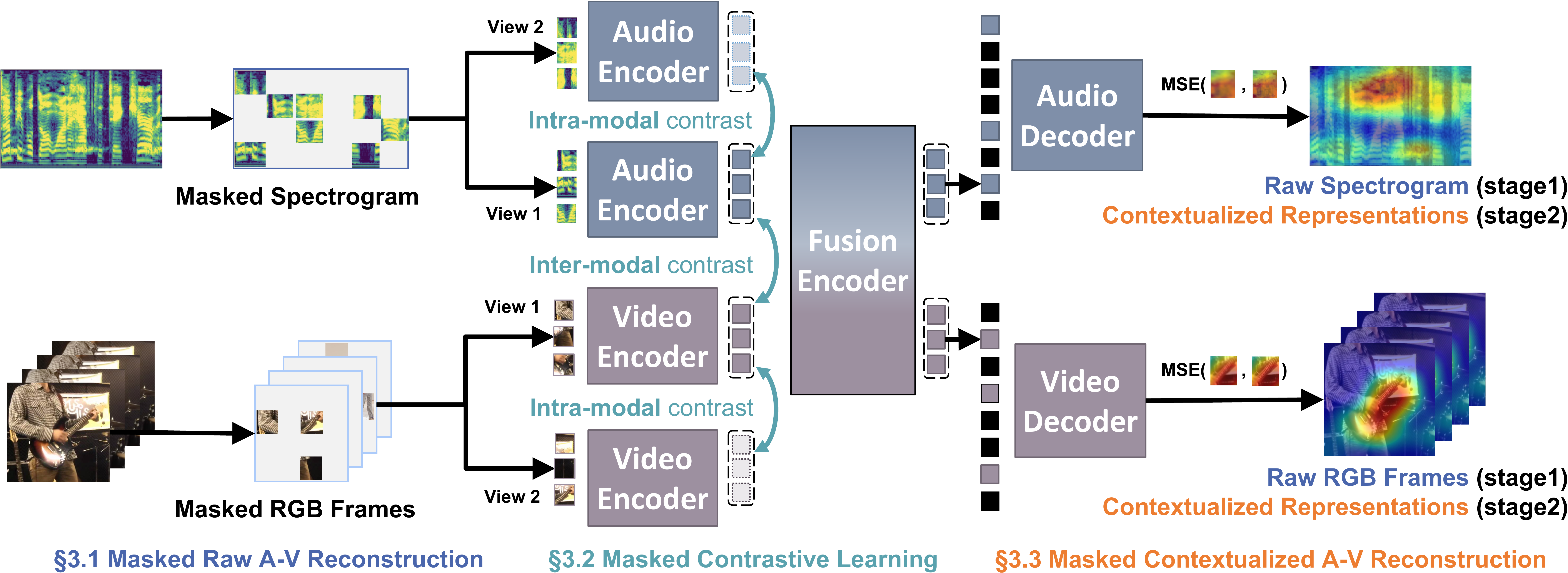}
    \caption{\textbf{Masked Audio-Video Learners (\ourmodel)} exploit three objectives for learning representations from audio video pairs with masking: (1) Raw audio-video reconstruction. (2) Inter-modal and intra-modal contrastive learning with masking. (3)  Reconstructing aligned and contextualized audio-video representations via student-teacher learning. (Please see Fig.~\ref{fig:mmkd} for details.) 
    \label{fig:overview}}
    \vspace{-1.5em}
\end{figure}

In this work, we develop a unified approach that combines MAE and contrastive learning to achieve effective and efficient audio-video representation learning for both audio-video tasks and audio-only tasks.
This combination presents challenges as it requires careful model design to handle heterogeneous multi-modal inputs in MAE and adapt contrastive learning with the majority of inputs masked for efficiency. Our approach, called Masked Audio-Video Learners (\ourmodel) (see Fig.~\ref{fig:overview}), consists of a pair of audio-video encoders, a fusion-encoder, and separate decoders for reconstructing raw or contextualized inputs. We design \ourmodel with three types of objectives outlined next. 


Firstly, we extend uni-modal MAE to multimodal and utilize a fusion encoder to exchange information from all modalities. \ourmodel reconstructs raw inputs that has been removed under a high masking ratio (\eg 80\%). With this, it learns \emph{complementary} audio-video representations by reconstructing a single modality input, with the supplementary context from the other.
Secondly, we adapt contrastive learning~\cite{nce,info_nce}, under a high masking ratio, to learn an \emph{aligned} audio-video latent space efficiently.
\ourmodel employs two types contrastive objectives:
(i) An \emph{inter-modal} contrast that brings together paired video and audio clips from the same video and contrasts them with other samples.
(ii) An \emph{intra-modal} contrast that draws closer the two masked views of the same audio or video, while pushing away other samples from the same modality.

Thirdly, 
unlike conventional MAEs that focus on reconstructing \emph{heterogeneous} and \emph{raw} inputs (audio \textit{or} video), we propose a novel pre-training task that reconstructs \emph{homogeneous} and \emph{contextualized} audio-video representations in a joint (audio-\textit{and}-video) latent space (illustrated in Fig.~\ref{fig:mmkd}).

We are motivated by the recent successes in visual SSL that use disparate teacher models to generate contextualized representations as reconstruction targets.
For example, BEiT~\cite{beit} uses features by DALL-E~\cite{dalle} as its target.
\mbox{MaskFeat}~\cite{maskfeat} predicts masked target features such as HOG~\cite{Dalal2005} or DINO \cite{dino}. 
Inspired by these uni-modal teachers, we take a step further to propose a new masked \emph{contextualized audio-video reconstruction} pretext task.
Our core idea is: instead of predicting raw inputs in heterogeneous modalities, the model with masked-view inputs jointly predicts contextualized audio-video representations in the homogeneous (aligned) latent space generated by a teacher (\ie an identical model pre-trained with the masked reconstruction and contrastive objectives above) with complete-view inputs.
This approach ensures the students learn well-aligned and contextualized audio-video representations and thus improves its performance on downstream tasks.
To achieve this, without relying on external teacher models, 
we propose a simple two-stage self-training framework (illustrated in Fig.~\ref{fig:mmkd}). 
In stage-1, we train the teacher \ourmodel and use raw inputs as its reconstruction targets. 
In stage-2, the student \ourmodel (the final model for later fine-tuning) learns to reconstruct the aligned and contextualized audio-video representations generated by the \ourmodel teacher.

Our experimental results confirm \ourmodel's superiority in audio-video classification (AudioSet-20K, AudioSet-2M, VGGSounnd) and retrieval tasks (VTT and YouCook) where it surpasses the best self-supervised and supervised pre-trained models by large margins. Notably, \ourmodel not only learns strong joint audio-video representations, but can also improve single modality encoders, without using the other modality it pre-trained with during fine-tuning or inference.

In summary, \ourmodel makes the following contributions for learning self-supervised audio-video representations: (1) Efficient contrastive learning for both intra-modal and inter-modal context using masking. (2) Introducing a new pretext task for multimodal MAE that predicts aligned and contextualized audio-video reconstruction, surpassing the performance of reconstructing uni-modal raw inputs. (3) Setting new state-of-the-art results in seven audio-visual classification and retrieval tasks and (4) audio-only tasks under the SSL setup without using labeled data for pre-training.

\vspace{-.75em}
\section{Related Work}\label{sec:related-work}
\vspace{-.75em}

\paragraph{Supervised Audio-Video Models.}
The connection between visual signals and co-occurring audio context for language acquisition and world comprehension in infants~\cite{roy1999learning} has motivated the research of audio-visual learning~\cite{ramachandram2017deep, chrupala2022visually}.
Previous studies have explored audio-visual ASR~\cite{chen1998audio, potamianos2003recent} and person identification \cite{aleksic2006audio} prior to the deep learning era. 
More recently, there has been significant attention on audio-video representation learning for classification \cite{ngiam2011multimodal, kim2013deep, ephrat2018looking, kazakos2019epic, avslowfast, nagrani2021attention}.
However, these supervised approaches rely on abundant labeled data, which are costly and time-consuming to obtain. 
While efforts have been made to create large-scale labeled datasets \cite{harwath2018jointly}, the dependency on extensive annotation hinders progress in audio-video modeling.
In contrast, \ourmodel focuses on self-supervised learning of robust audio-video representations without the need for labeled data.

\paragraph{Self-Supervised Audio-Video Representation Learning.}
To exploit abundant unlabeled video and audio content on the Internet and reduce annotation efforts, self-supervised techniques have been explored for learning audio-video representations~\cite{aytar2016soundnet, arandjelovic2017look, arandjelovic2018objects, korbar2018cooperative, shi2022learning}.
Inter-modal contrastive learning is a widely used approach that learns to associate paired audio-video clips as self-supervision~\cite{ma2020active, owens2018audio, morgado2021audio, stica}.
Techniques such as data augmentation~\cite{patrick2021compositions, wang2021multimodal} and harder negatives mining~\cite{zeng2021contrastive, recasens2021broaden, morgado2021robust} have been studied to improve its performance.
\ourmodel unifies masked autoencoding and contrastive learning. We notice a concurrent and independent study CAV-MAE~\cite{cavmae} uses inter-modal contrastive objective and MAE~\cite{mae} to reconstruct raw inputs.
Unlike their approach, \ourmodel takes a step further to incorporate intra-modal contrastive learning from two masked views. Most importantly, \ourmodel employs student-teacher learning to reconstruct contextualized and aligned representations instead of raw inputs as in CAV-MAE. These innovations lead to superior performance.

\paragraph{Student-teacher learning}
or knowledge distillation (KD)~\cite{hinton2015distilling, tian2019contrastive, park2019relational} was originally developed for model compression~\cite{ mishra2018apprentice, cho2019efficacy}, aiming to transfer knowledge from a larger teacher model to a smaller student model.
In the context of SSL, KD has recently gained attention for improving the distribution of representations by distilling~\cite{wei2022contrastive} or by reconstructing contextualized targets from a teacher model~\cite{byol, beit, mvp}.
Approaches like MoCo~\cite{moco}, DINO~\cite{dino}, and dBOT~\cite{liu2022exploring} utilize self-training to bootstrap targets from previous model snapshots during pre-training.
Data2vec~\cite{d2v} performs self-training in disparate modalities, where each one bootstraps complete-view contextualized targets independently. 
In contrast, to our best knowledge, \ourmodel is the first work that employs masked multimodal context for self-training. It uniquely incorporates complete-view multimodal fusion in the teacher model, while the student model receives masked-view inputs.

\section{Masked Audio-Video Learners\label{sec:avmae}}
We introduce Masked Audio-Video Learners (\ourmodel), a self-supervised audio-video representation learning framework (see Fig.~\ref{fig:overview}). 
\ourmodel consists of two stages:
In stage-1, 
\ourmodel jointly reconstructs raw spectrograms and RGB frames by exploiting complementary information from each modality (\S\ref{sec:maskedav}), and couples this with contrastive learning to encourage alignment between semantically similar instances, both \emph{within} and \emph{across} modalities. (\S\ref{sec:contra}). 
In stage-2 (see Fig.~\ref{fig:mmkd}),
we employ either the stage-1 model trained with raw targets (in iter. 1) or the last trained stage-2 model (in iter. 2+) as the teacher for self-training. We use the teacher's aligned and contextualized audio-video representations, obtained with complete-view inputs, to guide the student with masked-view inputs (\S\ref{sec:avmaet}). 
In the following, we provide details of these three types of self-supervision, starting with \ourmodel-stage1 trained with raw inputs.

\subsection{Masked Raw Audio-Video Reconstruction\label{sec:maskedav}}

Human perception involves processing visual and acoustic context jointly. In line with this, \ourmodel utilizes multimodal Transformers to fuse and exploit the complementary information from both audio and video. 
It aims to reconstruct audio and video simultaneously as self-supervision, which sets it apart from uni-modal MAE approaches such as MAE~\cite{mae}, Audio-MAE~\cite{huang2022masked}, or Video-MAE~\cite{feichtenhofer2022masked}.

Given a raw audio-video pair $(\mathrm{a}, \mathrm{v})\in\mathcal{D}$,
we begin by patchifying and tokenizing raw audio spectrograms and video frames into audio and video tokens.
This involves applying (audio/video) transforms, followed by 2D/3D-convolutions and flattening. This process embeds raw inputs into $\mathbf{a}=[a_1 \dots a_N]$ audio spectrogram tokens and $\mathbf{v}=[v_1 \dots v_M]$ video tokens, where $a_i, v_j \in\mathbb{R}^H$. To incorporate positional information, similar to MAE, we employ fixed 2D sinusoidal positional embeddings with the embedded tokens for each modality.
We then randomly mask the majority (\ie 80\%) of audio and video tokens. Only the remaining $20\%$ unmasked audio ($\mathbf{a}'$) and video ($\mathbf{v}'$) tokens are respectively fed into the audio ($f_\text{a}(.)$) and video ($f_\text{v}(.)$) Transformer encoders. 
This process results in uni-modal embeddings, denoted as $\mathbf{a}_\text{um}=f_\text{a}(\mathbf{a}')$ and $\mathbf{v}_\text{um}=f_\text{v}(\mathbf{v}')$. 

Following the uni-modal encoders, we incorporate a multimodal \textit{fusion} encoder denoted as $g_{\text{av}}(.)$ to model multimodal context. 
We explore two variants for this purpose: Vanilla Transformers~\cite{vaswani2017attention} and  Multimodal Bottleneck Transformers (MBT)~\cite{nagrani2021attention}. 
For vanilla Transformers, we jointly encode the audio and video tokens by: $(\mathbf{a}_{\text{um}}^{l+1}\|\mathbf{v}_{\text{um}}^{l+1})=\text{Transformer}^{l}(\mathbf{a}_{\text{um}}^l\|\mathbf{v_{\text{um}}}^l)$, where ``$\|$'' represents concatenation. Details of MBT implementation is in Appendix.
In both variants, we stack $L$ Transformer layers to obtain the jointly encoded top-layer outputs $\mathbf{a}_{\text{mm}}$ and $\mathbf{v}_{\text{mm}}$.

For reconstruction,
we employ vanilla Transformer blocks as the audio $f_\text{a}^{-1}(.)$ and video $f_\text{v}^{-1}(.)$ decoders. The fusion encoder's outputs ($\mathbf{a}_{\text{mm}}$ and $\mathbf{v}_{\text{mm}}$) are firstly projected and padded with trainable \texttt{[MASK]} tokens. After restoring the original order (time-frequency for audio and space-time for video tokens), we add the decoders' (fixed 2-D sinusoidal) positional embeddings and input the restored sequences into the decoders.
At the top of the decoders, we incorporate linear heads to reconstruct the raw inputs.
Specifically, the decoder outputs for spectrogram reconstruction are denoted as $\mathbf{\hat{a}}=f_{\text{a}}^{-1}(g_{\text{av}}(f_{\text{a}}(\mathbf{a}')))$ and for RGB frame reconstruction as $\mathbf{\hat{v}}=f_{\text{v}}^{-1}(g_{\text{av}}(f_{\text{v}}(\mathbf{v}')))$. For notation clarity, we omit the \texttt{[MASK]} tokens and linear projection head.
Let $\hat{\mathbf{a}}_i,\mathbf{a}_i^{\text{raw}} \in \mathbb{R}^{H_{\text{raw}}^{\text{a}}}; i=1\dots n$ denote the audio decoder's output and the ground truth reference of the $i$-th masked spectrogram patch. Similarly, $\hat{\mathbf{v}}_j,\mathbf{v}_j^{\text{raw}} \in \mathbb{R}^{H_{\text{raw}}^{\text{v}}}; j=1\dots m$ for video patches.
In masked audio-video reconstruction, \ourmodel is self-supervised by minimizing the mean squared error (MSE) loss $\mathcal{L}_r^{\text{raw}}$ defined as:
\begin{equation}
    \mathcal{L}_r^{\text{raw}} = \frac{1}{n}\sum_{i=1}^n(\hat{\mathbf{a}}_i-\mathbf{a}_i^{\text{raw}})^2 + \frac{1}{m}\sum_{j=1}^m(\hat{\mathbf{v}}_j-\mathbf{v}_j^{\text{raw}})^2.\label{equ:raw}
\end{equation}

\vspace{-0.5em}
\subsection{Contrastive Audio-Video Learning with Masking}\label{sec:contra}


Contrastive learning is widely used to learn uni-modal~\cite{simclr,moco,mocov3} and multimodal~\cite{clip,stica} representations by aligning multiple ``views'' of the same instance.
These views can be either \emph{within}-modality observations of the instance itself (e.g.,  the same audio under different volumes) or semantic observations \emph{across} modalities (e.g., a video and its corresponding audio).
\ourmodel utilizes InfoNCE~\cite{info_nce} loss for contrastive learning.
Let $\mathbf{x}=[\mathbf{x}_1\dots\mathbf{x}_B],\mathbf{y}=[\mathbf{y}_1\dots\mathbf{y}_B]; \mathbf{x}_i, \mathbf{y}_j\in\mathbb{R}^{H}$ be 
the instance-level representations of audio/video in a batch of size $B$.  
The contrastive loss $\mathcal{L}_{\text{c}}(\mathbf{x},\mathbf{y})$ is defined as:
\begin{equation}
\label{equ:contrastive}
\mathcal{L}_{\text{c}}(\mathbf{x},\mathbf{y}) = 
-\frac{1}{B} \sum_{i=1}^B {\rm log}  
\frac{ {\rm exp} (\text{S}(\mathbf{x}_i,\mathbf{y}_i)/\tau)}
{\sum_{j=1}^{B} {\rm exp} (\text{S}(\mathbf{x}_i,\mathbf{y}_j)/\tau)) },
\end{equation}
where $\text{S}(\mathbf{x}_i,\mathbf{y}_j)=\frac{\mathbf{x}_i^T\mathbf{y}_j}{\|\mathbf{x}_i\| \|\mathbf{y}_j\|}$ is the cosine similarity between $\mathbf{x}_i$, $\mathbf{y}_j$ and $\tau$ is the softmax temperature. 
\ourmodel employs two types of contrastive losses for self-supervision listed below:


\noindent\textbf{Inter-modal contrastive learning} facilitates alignments \emph{across} modalities.
We first average the sequence of uni-modal encoder outputs\footnote{We do not use the fusion output $\mathbf{a}_{\text{mm}}$ and $\mathbf{v}_{\text{mm}}$ for contrastive learning since the fusion layer provides a shortcut/leakage of information exchange for every paired audio-video clip, resulting in sub-optimal performance.} as the instance-level representations, namely, 
$\mathbf{a}_{\text{emb}}=\text{Avg}(\mathbf{a}_{\text{um}})$ and $\mathbf{v}_{\text{emb}}=\text{Avg}(\mathbf{v}_{\text{um}})$. 
The positive pairs (the numerator in Eq.\eqref{equ:contrastive}) consist of video and audio clips from the same video, while all the other combinations of sampled audio-video pairs are considered negatives (the denominator).
Inter-modal contrast encourages the representations of paired audio and video to be closer to each other, while simultaneously pushing away mismatched ones.

\noindent\textbf{Intra-modal contrastive learning} promotes alignment \emph{within} each modality. It aims to bring the representations of different views, such as different augmented versions of an audio (or video), closer to each other.
To achieve this, for each modality, we apply random masking and sample a second view that contains 20\% tokens for encoding. 
The idea is to treat masking as a form of augmentation to generate two contrasting views in the same modality, which has been proven effective in visual SSL~\cite{moco,simclr}. 
The two views of the same instance are considered as a positive pair, which are then contrasted against the other (negative) combinations of the instances in the same modality. 
Formally, let $\mathbf{\bar{a}}_{\text{emb}}$ and $\mathbf{\bar{v}}_{\text{emb}}$ be the embeddings of the second-view audio and video. \ourmodel employs the inter-modal ($\mathcal{L}_{\text{c}}^{\text{inter}}$) and intra-modal ($\mathcal{L}_{\text{c}}^{\text{intra}}$) contrastive  objectives defined as:
\begin{equation}\label{eq:contrastive}
\mathcal{L}_{\text{c}}^{\text{inter}}=\frac{1}{2}
\left[\mathcal{L}_c(\mathbf{a}_{\text{emb}},\mathbf{v}_{\text{emb}})+\mathcal{L}_c(\mathbf{v}_{\text{emb}},\mathbf{a}_{\text{emb}})
\right],
\quad
\mathcal{L}_{\text{c}}^{\text{intra}}=\frac{1}{2}\left[
\mathcal{L}_c(\mathbf{a}_{\text{emb}},\mathbf{\bar{a}}_{\text{emb}})+
\mathcal{L}_c(\mathbf{v}_{\text{emb}},\mathbf{\bar{v}}_{\text{emb}})
\right],
\end{equation}

These contrastive losses effectively learn a latent audio-video space where semantically similar instances are close to each other.
Note that unlike prior work (\eg~\cite{moco, simclr}), \ourmodel performs contrastive learning under masked-view. This leads to improved computation efficiency as only a small portion of input tokens are encoded for contrast. And different from CAV-MAE~\cite{cavmae}, \ourmodel additionally incorporates intra-modal contrast, which yields superior performance. Overall, 
let $\alpha$ and $\beta$ be the weights that balance loss terms, \ourmodel-stage1 is self-supervised by minimizing:
\begin{equation}
\mathcal{L}_{\text{\ourmodel}} = \mathcal{L}_{r}^{\text{raw}} + \alpha \mathcal{L}_{\text{c}}^{\text{inter}} + \beta \mathcal{L}_{\text{c}}^{\text{intra}} ,
\label{eq:1st_avmae}
\end{equation}

\begin{figure*}[t!]
    \centering
    \includegraphics[width=1.0\linewidth]{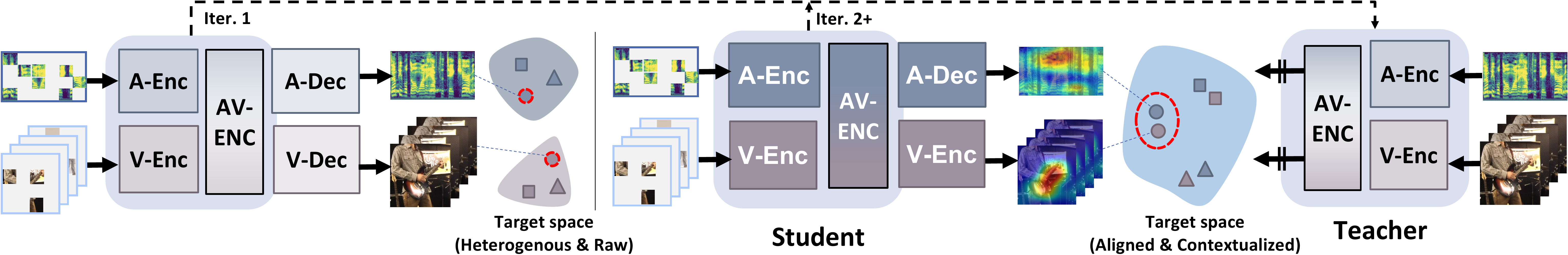} 
    \vspace{-10pt}
    \caption{\textbf{Masked contextualized audio-video reconstruction} in the joint latent space. Stage-1 (left): Training \ourmodel with raw inputs as targets. Stage-2 (right): Self-training \ourmodel student by reconstructing \ourmodel teacher's aligned and contextualized audio-video representations generated with complete inputs.
    Repeat stage-2 for $K$ iterations. In the first iteration of stage-2, the stage-1 model is used as the teacher. In subsequent iterations (iteration 2+), the last trained stage-2 student model serves as the new teacher.\label{fig:mmkd}}
    \vspace{-10pt}
\end{figure*}
\subsection{Masked Contextualized Audio-Video Reconstruction}\label{sec:avmaet}

To learn robust audio-video representations, we go beyond raw input reconstruction in uni-model MAEs~\cite{mae,huang2022masked,feichtenhofer2016spatiotemporal}, multimodal MAE~\cite{bachmann2022multimae}, and CAV-MAE~\cite{cavmae}.
We propose a new pretext task that reconstructs  contextualized audio-video representations in the joint latent space.
To achieve this, 
we employ a simple two-stage training framework illustrated in Fig.~\ref{fig:mmkd}.
In stage-1, we train \ourmodel with Eq.\eqref{eq:1st_avmae} to reconstruct raw inputs.
In stage-2, we employ a student-teacher learning framework for iterative self-training.
In the first iteration, the pre-trained \ourmodel from stage-1 is frozen and serves as the \emph{teacher} model. 
It generates audio-video representations with complete-view inputs, which are used as the reconstruction targets to guide the re-initialized \emph{student} \ourmodel.
In the subsequent iterations, the last trained stage-2 student \ourmodel serves as the new teacher.

Formally, we provide the teacher model's encoders with complete-view audio and video inputs to generate aligned and contextualized targets: $\mathbf{a}^{\text{Teacher}}\|\mathbf{v}^{\text{Teacher}}=g_{\text{av}}^{\text{Teacher}}(f_\text{a}^{\text{Teacher}}(\mathbf{a})\|f_\text{v}^{\text{Teacher}}(\mathbf{v}))$.
The student \ourmodel then learns to reconstruct these contextualized targets with the masked-view inputs. Precisely, $\mathbf{\tilde{a}}=f_{\text{a}}^{-1}(g_{\text{av}}(f_{\text{a}}(\mathbf{a}')))$ for audio and $\mathbf{\tilde{v}}=f_{\text{v}}^{-1}(g_{\text{av}}(f_{\text{v}}(\mathbf{v}')))$ for video, where $\mathbf{a}^{\text{Teacher}}, \mathbf{v}^{\text{Teacher}}, \mathbf{\tilde{a}}, \mathbf{\tilde{v}} \in\mathbb{R}^{H}$.
The contextualized reconstruction objective is defined as:
\begin{equation}\label{eq:loss:context}
    \mathcal{L}_r^{\text{context}} = \frac{1}{n}\sum_{i=1}^n(\tilde{\mathbf{a}}_i-\mathbf{a}_i^{\text{Teacher}})^2 + \frac{1}{m}\sum_{j=1}^m(\tilde{\mathbf{v}}_j-\mathbf{v}_j^{\text{Teacher}})^2.
\end{equation}

In the stage-2 training, we jointly minimize the masked contextualized reconstruction loss and the contrastive loss. The stage-2 (student) \ourmodel's objective  is:
\begin{equation}
\mathcal{L}_{\text{\ourmodel}} = \mathcal{L}_{r}^{\text{context}} + \alpha \mathcal{L}_{\text{c}}^{\text{inter}} + \beta \mathcal{L}_{\text{c}}^{\text{intra}} .
\label{eq:2nd_avmae}
\end{equation}
Note that Eq.\eqref{eq:2nd_avmae} contains pure latent targets\footnote{Empirically, we found no gain when multi-tasking with stage-1 autoencoding loss on raw pixel/spectrogram.}. 
After pre-training, we then fine-tune the audio/video encoders in the final stage-2 \ourmodel student model in the downstream tasks.

\section{Experiments\label{sec:exp}}
We performed comprehensive evaluations, including audio-video classification tasks on AudioSet~\cite{gemmeke2017audio} (AS-2M and AS-20K), and VGGSound~\cite{chen2020vggsound}.
Also, we conducted audio-to-video retrieval experiments on MSR-VTT~\cite{xu2016msr} and YouCook~\cite{zhou2017towards}. We use AS-20K for model analysis and ablation studies.

\subsection{Datasets}
\textbf{AudioSet} contains 2 million 10-second YouTube clips for audio event detection.
527 event types are weakly labeled~\cite{tagging_right,vggish,hershey2021benefit} for each clip, and multiple events can occur in one clip.
AudioSet's full training set has two subsets: A class-wise \emph{balanced} (22K clips) and an \emph{unbalanced} (2M clips) set.
The \emph{eval} set has 20K clips. 
We downloaded 1.97M unbalanced training, 20K balanced training, and 19K evaluation clips.
We use the full (unbalanced+balanced) training set for pre-training.
In the AS-2M task, we fine-tune on the full training set.
In the AS-20K task, we fine-tune only on the 20K balanced training set.
We report the classification mAP on the 19K \emph{eval} set used by AST~\cite{gong21b_interspeech}.

\textbf{VGGSound} comprises approximately 200K 10-second video clips annotated with 309 event types that include human actions, sound-emitting objects, \etc. Unlike AudioSet, VGGSound ensures that an audio event is also visually present in its corresponding clips. VGGSound is divided into 183K training and 15K testing samples.  We report top-1 testing classification accuracy.

\subsection{Implementation Details}
\label{sec:imp}

\ourmodel adopts different temporal footprints for audio and video.
For audio, following~\cite{nagrani2021attention,gong21b_interspeech}, it transforms a 10-second audio under 16K sampling rate into 128 Mel-frequency bands with a 25ms Hanning window shifting every 10ms. The resulting spectrogram is of dimension $1024\times128$.
\ourmodel then tokenizes it into non-overlapping $16\times16$ patches where both time and frequency have a kernel and stride of 16. The flattened audio tokens have a sequence length $N$ of 512.
For video, \ourmodel processes 4-second clips under 2 frames per second. Each frame has a size of $224\times224$. 
Tokenization is achieved by a 3D convolution, where the spatial kernel and stride are 16, and the temporal kernel and stride are 2.
The flattened video tokens have a sequence length $M$ of 784.

Following the design choices in MAE~\cite{mae}, \ourmodel employs 12-layer Transformers (ViT-B) with 12 attention heads as the encoders for each modality .  
The embedding dimension $H$ is set to $768$.
The audio-video fusion encoder layer consists of a 2-layer Transformer (vanilla or MBT) on top of the uni-modal encoders.  
Similarly, the audio and video decoders utilize 8-layer Transformers with an embedding dimension of 512 and 16 attention heads. \ourmodel's audio/video encoder and decoder have 86M and 27M parameters, respectively. The floating point operations (FLOPs) for the audio encoder are 48.6G, comparable to the audio encoders in Audio-MAE~\cite{huang2022masked} and CAV-MAE~\cite{cavmae}.

For pre-training \ourmodel's audio branch, we randomly initialize it from scratch.
For the visual branch, we either randomly initialize it or initialize it with the self-supervised MAE~\cite{mae} pre-trained on ImageNet (compared in Table~\ref{tab:as_result}).
Notably, \ourmodel operates under the fully \emph{self-supervised} setup.

\subsection{Experimental Setup}
We pre-train \ourmodel on AS-2M without using any of AS-2M labels. We use 80\% masking for audio and video. For balancing the losses in Eq.\eqref{eq:2nd_avmae}, we set $\alpha=0.1, \tau_{\text{c}}^{\text{inter}}=0.1$ and $\beta=0.01, \tau_{\text{c}}^{\text{intra}}=1.0$. These hyperparameters scale the gradients from the three losses into a comparable range to improve training stability. We pre-train with 64 GPUs with a 512 accumulated batch size and a 0.0002 learning rate. We pre-train for 20 epochs in stage-1 and in each iteration of stage-2 (for  ($K=3$) iterations). Each session takes 20 hours, resulting in a total pre-training time around 80 hours.

We then evaluate audio and video representations by fine-tuning the pre-trained \ourmodel in audio-video classification tasks on AS-20K, AS-20M, and VGGSound. We test 3 scenarios: (1) audio-only, (2) video-only, and (3) audio+video. 
Following MAE~\cite{mae}, we only keep the pre-trained encoders and use the average-pooled top-layer outputs for classification.
We adopt the standard fine-tuning pipeline and augmentation in prior audio/audio-video classification works~\cite{gong21b_interspeech, huang2022masked, nagrani2021attention}. Specifically, we employ SpecAug~\cite{Park2019SpecAugmentAS}, mixup~\cite{mixup}, balanced sampling~\cite{tagging_right}, and fine-tuning masking~\cite{huang2022masked}. 
For video, we use standard video augmentations used in video classification~\cite{maskedfeat, feichtenhofer2022masked}.
To address the discrepancy in convergence rate between audio and video, we apply a 50\% learning rate reduction for the video encoder during the audio+video (fusion) fine-tuning. 
Fine-tuning for 100 epochs on AS-2M takes 10 hours. 
Please refer to Appendix for additional implementation and experimental details.

In the following, we first use AS-20K to analyze \ourmodel's performance in \S\ref{sec:ablation}. We then present the main comparison to prior works in~\S\ref{sec:sota}. The \colorbox{gray!30}{gray} entries represent the shared default setup.

\subsection{Model Analysis} \label{sec:ablation}

\paragraph{Masked Raw Audio-Video Reconstruction\label{sec:exp:masked_av_learning}.}
Table~\ref{tab:joint_encoder} shows the contribution of \ourmodel's audio-video fusion encoder layer, which exploits complementary multimodal information for reconstructing raw inputs in stage-1.
The `None'' (col. 1) indicates the uni-modal MAE baselines without the fusion layer, namely Audio-MAE and Video-MAE under our implementation. 
The fusion layer, whether it's a vanilla Transformer or MBT, contributes up to 0.4 mAP gains (col. 2-5). 
While the MBT layers show better results, the improvements is not significant compared to using vanilla Transformers.
Also, increasing the depth of the fusion encoder only leads to marginal gains.
For simplicity, we default to using a 2-layer vanilla Transformer as the fusion encoder.

\begin{table}[th]
    \centering
    \parbox{.55\linewidth}{
        \centering\footnotesize
        \begin{tabular}{c|x{25}>{\columncolor{defaultcolor}}x{30}x{30}x{35}}
        Fusion 
        & None & Vanilla\textsuperscript{(2)} & Vanilla\textsuperscript{(4)} & MBT\textsuperscript{(4)}\\
        \shline
        Audio & 36.4 & 36.8\textsubscript{\tiny(+0.4)} & 36.7\textsubscript{\tiny(+0.3)} & 36.8\textsubscript{\tiny(+0.4)} \\
        Video & 17.4 & 17.7\textsubscript{\tiny(+0.3)} & 17.6\textsubscript{\tiny(+0.2)} & 17.7\textsubscript{\tiny(+0.3)}
        \end{tabular}
        \vspace{4pt}
        \caption{\textbf{Fusion Encoder}: Transformer type\textsuperscript{(\# layers)}\label{tab:joint_encoder}. Note that the `None' are the uni-modal MAE baselines (Audio-MAE and Video-MAE).}
    }    
    \hfill
    \parbox{.41\linewidth}{
        \centering\footnotesize
        \setlength\tabcolsep{3.0pt}
        \begin{tabular}{c|x{20}x{20}x{20}>{\columncolor{defaultcolor}}x{20}x{20}}
        Ratio &  40\% & 60\% & 70\% & 80\% & 90\%  \\
        \shline
        Audio &  35.6 & 36.5 & 36.7 & 36.8 & 36.8\\
        Video &  16.8 & 17.3 & 17.5 & 17.7 & 17.5 \\
        \end{tabular}
        \vspace{4pt}
        \caption{\textbf{Masking Ratio}. mAP$\uparrow$ in \mbox{AS-20K} for individual fine-tuning as audio or video classification.\label{tab:mask_ratio}}
    }
    \vspace{-1em}
\end{table}

Subsequently, we investigate the masking ratio, a key hyperparameter in masked autoencoding. We use the same masking ratio for audio and video, and ablate different values in Table~\ref{tab:mask_ratio}. The results suggest a masking ratio of 80\% achieves the best performance. This aligns with the optimal values of 80\% in Audio-MAE~\cite{huang2022masked} and 90\% in Video-MAE~\cite{feichtenhofer2022masked}. \ourmodel encodes only the non-masked tokens significantly reduces the sequence length, resulting in more efficient computation, as the complexity of self-attention layers in Transformers scales quadratically with the sequence length.

\paragraph{Contrastive Audio-Video Learning with Masking.}
Next, we examine the benefits when contrastive learning is used in conjunction with MAE. \ourmodel performs efficient contrastive learning with only 20\% of visible patches in each modality (80\% are masked). 
The results in Table~\ref{tab:contrastive} demonstrate the significance of contrastive losses for learning audio-video representations, even with only 20\% visible tokens.
Unlike CAV-MAE~\cite{cavmae} which uses only inter-modal contrast, we observe that both inter-modal and intra-modal contrast are essential, and combining them delivers the best performance.

\begin{table}[!h]
\centering
\setlength\tabcolsep{3.0pt}
\footnotesize
\begin{tabular}{c|x{25}x{30}x{30}>{\columncolor{defaultcolor}}x{40}c}
Objs. 
& None & Inter & Intra & Inter+intra \\
\shline
Audio & 36.8 & 38.4\textsubscript{\tiny(+1.6)} & 38.1\textsubscript{\tiny(+1.3)} & 39.0\textsubscript{\tiny(+2.2)} \\
Video & 17.7 & 21.0\textsubscript{\tiny(+3.3)} & 19.8\textsubscript{\tiny(+2.1)} & 22.2\textsubscript{\tiny(+4.5)}  \\
\end{tabular}
\vspace{5pt}
\caption{\textbf{Masked Contrastive Audio-Video Learning}\label{tab:contrastive}}
\vspace{-1em}
\end{table}

\paragraph{Masked Contextualized Audio-Video Reconstruction.}
As illustrated in Fig.~\ref{fig:mmkd}, \ourmodel employs a two-stage training. 
Following stage-1 training (Table~\ref{tab:contrastive}), \ourmodel's encoders acquire complementary audio-video representations through masked raw audio-video reconstruction, as well as aligned representations through masked contrastive learning.
In stage-2, we employ an iterative approach to initialize the student \ourmodel and train it under masked-views to predict the aligned and contextualized representations generated by the frozen \ourmodel teacher with full view. The teacher can be either from the stage-1 model (iter. 1) or the last trained stage-2 model (iter. 2 and 3).

\begin{table}[h]
    \begin{subtable}[h]{0.55\textwidth}
        \centering\footnotesize
        \setlength\tabcolsep{4.5pt}
        \begin{tabular}{c|ccc>{\columncolor{defaultcolor}}c}
             Target
             & Raw 
             & A/V-MAE
             & \makecell{M-Uni} 
             & \makecell{M-Fusion} 
            \\
            \shline
            Audio & 39.0 &  39.5\textsubscript{(\tiny+0.5)} & 40.2\textsubscript{(\tiny+1.2)} & 40.7\textsubscript{(\tiny+1.7)} \\
            Video & 22.2 &  23.1\textsubscript{(\tiny+0.9)} & 23.8\textsubscript{(\tiny+1.6)} & 24.1\textsubscript{(\tiny+1.9)} \\
        \end{tabular}
        \caption{\textbf{Reconstruction Targets} (Stage-2, iteration-1)\label{tab:target}}
    \end{subtable}
    \begin{subtable}[h]{0.42\textwidth}
        \centering\footnotesize
        \setlength\tabcolsep{3.0pt}
        \begin{tabular}{c|ccc>{\columncolor{defaultcolor}}c}
             Iter.
             & None
             & 1 
             & 2
             & 3
            \\
            \shline
            Audio & 39.0 & 40.7\textsubscript{(\tiny+1.7)} & 41.5\textsubscript{(\tiny+2.5)} & 41.8\textsubscript{(\tiny+2.8)} \\
            Video & 22.2 & 24.1\textsubscript{(\tiny+1.9)} & 24.6\textsubscript{(\tiny+2.4)} & 24.8\textsubscript{(\tiny+2.6)} \\
        \end{tabular}    
        \caption{\textbf{Iterations} (Stage-2)\label{tab:iteration}}
    \end{subtable}
  \caption{\textbf{Targets} and \textbf{iterations} in \ourmodel's stage-2 contextualized audio-video reconstruction. (a) Target comparison in the 1\textsuperscript{st} iteration. Raw: spectrogram/RGB frames. Contextualized: A-MAE/V-MAE (uni-modal MAE output); M-Uni (\ourmodel's uni-modal output); M-Fusion (\ourmodel's multimodal fusion output). (b) Iteration comparison. `None' is the \ourmodel-stage1 baseline in Table~\ref{tab:contrastive}.}
\end{table}

We first compare the reconstruction targets in Table~\ref{tab:target} (stage-2, iter. 1). By default \ourmodel student predicts the last trained \ourmodel teacher's multimodal fusion encoder outputs (\ie, $g_{\text{av}}^{\text{Teacher}}( f_\text{a}^{\text{Teacher}}(\mathbf{a})\|f_\text{v}^{\text{Teacher}}(\mathbf{v}))$, M-Fusion, col. 4).
The results show that this improves over predicting (col. 1) raw spectrogram/RGB frames (+1.7 audio mAP) and (col. 2) separately pre-trained uni-modal Audio/Video-MAE's outputs (+1.2 audio mAP). 
The fused multimodal targets (M-Fusion, col. 4) are better (+0.5 audio mAP) than the uni-modal targets (M-Uni, col. 3) (\ie, $f_\text{a}^{\text{Teacher}}(\mathbf{a})$ and $f_\text{v}^{\text{Teacher}}(\mathbf{v})$ before the fusion encoder).

During the 2\textsuperscript{nd} and 3\textsuperscript{rd} iterations of stage-2 training, we leverage the last trained \ourmodel student as the new teacher. As shown in Table~\ref{tab:iteration}, substantial improvements are achieved over iterations through masked prediction of aligned and contextualized audio-video representations by the \ourmodel teacher. 
The accumulated gains with the proposed contextualized audio-video reconstruction in stage-2 are +2.8 audio and +2.6 video mAP \emph{without} using additional data or external teacher models.

\paragraph{Additional Ablations.}
In Table~\ref{tab:ablation} we ablate several more aspects of MAViL:
(a) Pre-training with more data is useful (100\% > 50\% > 10\% of AS-2M) (Tab.~\ref{tab:ablation:data_size}).
(b) ImageNet (IN) self-supervised pre-training is beneficial for the video encoder (Table~\ref{tab:ablation:backbone}). ImageNet supervised pre-training (IN-SL) is useful yet we avoid this to keep \ourmodel fully self-supervised. 
(c) Longer pre-training is beneficial and 20 epochs are sufficient (Table~\ref{tab:ablation:epoch}). 
(d) Increasing the encoder sizes (ViT-S/B/L) improves mAP (Table~\ref{tab:ablation:model_size}). We use ViT-B for efficiency and fair comparison as default.

\begin{table}[!h]
    \begin{minipage}{0.4\linewidth}  
        \centering
        \subfloat[\textbf{Dataset size}\label{tab:ablation:data_size}]{
            \centering\footnotesize
            \setlength\tabcolsep{1.5pt}
            \tablestyle{2pt}{1.00}
            \footnotesize
            \begin{tabular}{c|x{20}x{20}>{\columncolor{defaultcolor}}x{20}}
            \# & 200K & 1M & 2M \\
            \shline
            A & 34.1 & 40.5 & 41.8 \\
            V & 20.1 & 23.0 & 24.8 \\
            \end{tabular}     
        }
        \subfloat[\textbf{Visual backbone init.}\label{tab:ablation:backbone}]{
            \setlength\tabcolsep{1.5pt}
            \tablestyle{2pt}{1.00}
            \footnotesize
            \begin{tabular}{c|x{20}>{\columncolor{defaultcolor}}x{28}x{25}}
            Init. & Rand. & \small{IN-SSL} & \small{IN-SL} \\
            \shline
            A & 41.6 & 41.8 & 41.8 \\
            V & 23.7 & 24.8 & 26.1 \\
            \end{tabular}         
        }
        \hfill%
        \subfloat[\textbf{Pre-training epoch}\label{tab:ablation:epoch}]{
            \setlength\tabcolsep{1.5pt}
            \tablestyle{2pt}{1.00}
            \centering\footnotesize
            \begin{tabular}{c|x{18}>{\columncolor{defaultcolor}}x{18}x{18}}
            Ep. & 15 & 20 & 25 \\
            \shline
            A & 41.4 & 41.8 & 41.8 \\
            V & 23.5 & 24.8 & 24.7 \\
            \end{tabular} 
        }
        \subfloat[\textbf{Model size}\label{tab:ablation:model_size}]{
            \setlength\tabcolsep{1.5pt}
            \tablestyle{2pt}{1.00}
            \centering\footnotesize
            \begin{tabular}{c|x{24}>{\columncolor{defaultcolor}}x{24}x{24}}
            Sz. & ViT-S & \small{ViT-B} & ViT-L \\
            \shline
            A & 36.2 & 41.8 & 42.2 \\
            V & 20.5 & 24.8 & 27.0 \\
            \end{tabular} 
        }
        \hfill
    \end{minipage}
    \hspace{40pt}
    \begin{minipage}{0.4\linewidth}
    \centering
    \subfloat[\textbf{Module-wise contributions}\label{tab:ablation_all}]{
        \centering
        \setlength\tabcolsep{2.0pt}
        \footnotesize
        \begin{tabular}{lll}
        Method & Audio & Video \\
        \shline
        A-MAE/V-MAE (baseline) & 36.4 & 17.4 \\ \hline
        \multicolumn{3}{l}{\textit{\ourmodel stage-1}}\\
        + Joint AV-MAE & 36.8\textsubscript{(\tiny+0.4)} & 17.7\textsubscript{(\tiny+0.3)} \\ 
        + Inter contrast & 38.4 & 21.0 \\
        + Intra and Inter contrast & 39.0\textsubscript{(\tiny+2.2)} & 22.2\textsubscript{(\tiny+4.5)} \\ \hline
        \multicolumn{3}{l}{\textit{\ourmodel stage-2}}\\
        \rowcolor{gray!20} + Student-teacher learning & 41.8\textsubscript{(\tiny+2.8)} & 24.8\textsubscript{(\tiny+2.6)} \\        
        \end{tabular}
        }
    \vspace{5pt}
    \end{minipage}
    \caption{\textbf{Additional ablations} and \textbf{module-wise contributions} (mAP on AS-20K)}
    \label{tab:ablation}
    \vspace{-1em}
\end{table}

\paragraph{Summary.}
The module-wise contribution is summarized in Table~\ref{tab:ablation_all}. \ourmodel enhances uni-modal MAE performance by learning from audio \textit{and} video simultaneously, as evidenced by significant increases in audio mAP (36.4$\rightarrow$41.8, +5.4) and video mAP (17.4$\rightarrow$24.8, +7.4) on AS-20K.
It is noteworthy that while audio and video are jointly used in the pre-training phase, the fine-tuning is performed with uni-modal encoders. Therefore, the improvements observed in uni-modal classification showcase that learning from both modalities can enhance the performance of individual modalities.

\begin{table}[t]
\setlength\tabcolsep{5.0pt}
\centering
\footnotesize
\begin{tabular}{lccccccccccc}
 & & \multicolumn{3}{c}{AS-20K (mAP$\uparrow$)} & \multicolumn{3}{c}{AS-2M (mAP$\uparrow$)} & \multicolumn{3}{c}{VGGSound (Acc.$\uparrow$)} \\
\cmidrule(l){3-11} 
Method & PT & A & V & A+V & A & V & A+V & A & V & A+V  \\ \midrule
\multicolumn{8}{l}{\textit{\textbf{Audio-only Models}}}\\
Aud-SlowFast~\cite{Kazakos2021SlowFastAuditory} & - & - & - & - & - & - & - & 50.1 & - & - \\
VGGSound~\cite{chen2020vggsound} & - & - & - & - & - & - & - & 48.8 & - & - \\
PANNs~\cite{kong2020panns} & - & 27.8 & - & - & 43.9 & - & - & - & - & -  \\
AST~\cite{gong21b_interspeech} & IN-SL & 34.7 & - & - & 45.9 & - & - & - & - & - \\
HTS-AT~\cite{chen2022hts} & IN-SL & - & - & - & 47.1 & - & - & - & - & -  \\
PaSST~\cite{koutini2021efficient} & IN-SL & - & - & - & 47.1 & - & - & - & - & - \\
Data2vec~\cite{d2v} & AS-SSL & 34.5  & - & - & - & - & - & - & - & - \\ 
SS-AST~\cite{ssast} & AS-SSL & 31.0  & - & - & - & - & - & - & -& -  \\
MAE-AST~\cite{baade2022mae} & AS-SSL & 30.6  & - & - & - & - & - & - & -& -  \\
Aud-MAE~\cite{huang2022masked} & AS-SSL & 37.0  & - & - & 47.3 & - & - & - & - & - \\
\hline
\multicolumn{8}{l}{\textit{\textbf{Audio-Video Models}}}\\
G-Blend {\cite{wang2020makes}} & - & 29.1  & 22.1  & 37.8 & 32.4 & 18.8 & 41.8 & - & - & - \\
Perceiver {\cite{jaegle2021perceiver}} & - & - & - & - & 38.4 & 25.8 & 44.2 & - & - & - \\
Attn AV {\cite{fayek2020large}} & IN-SL & - & - & - & 38.4 & 25.7 & 44.2 & - & - & - \\
CAV-MAE {\cite{cavmae}} & IN-SSL, AS-SSL & 37.7 & 19.8  & 42.0 & 46.6 & 26.2 & 51.2 & 59.5 & 47.0 & 65.5 \\
\color{gray}MBT\textsuperscript{*}{\cite{nagrani2021attention}} & \color{gray}IN21K-SL & \color{gray}31.3 & \color{gray}27.7 & \color{gray}43.9 & \color{gray}41.5 & \color{gray}31.3 & \color{gray}49.6 & 52.3 & \textbf{51.2} & 64.1  \\      
\midrule
\ourmodel & AS-SSL & 41.6 & 23.7 & 44.6 & \textbf{48.7} & 28.3 & 51.9 & 60.6 & 50.0 & 66.5 \\ 

\rowcolor{gray!20} \ourmodel & IN-SSL, AS-SSL  & \textbf{41.8} & \textbf{24.8} & \textbf{44.9} & \textbf{48.7} & \textbf{30.3} & \textbf{53.3} & \textbf{60.8} & 50.9 & \textbf{67.1} \\
\end{tabular}
\vspace{5pt}
\caption{
\textbf{Comparison to prior work on AudioSet (AS-20K, AS-2M) and VGGSound} in the audio (A), video (V) and audio+video (A+V) classification tasks. PT: pre-training dataset and type; IN: ImageNet; SL: supervised learning; SSL: self-supervised learning; 
\textsuperscript{*}:We {\color{gray}de-emphasize} the model using non-standard dataset splits.
We bold the best-performing single model. 
\label{tab:as_result}}
\vspace{-1em}
\end{table}

\subsection{Main Results} \label{sec:sota}

The full model is the stage-2 \ourmodel student (ViT-B) trained with Eq.\eqref{eq:2nd_avmae} for 3 iterations. We quantitatively compare it with other previous models on the seven benchmark tasks.
Due to page limit, we include the additional experiments and the qualitative reconstructions by the stage-1 \ourmodel's audio/video decoders in Appendix.

\paragraph{Audio-Video Event Classification.}
Table~\ref{tab:as_result} presents a comparison of \ourmodel's audio (A), video (V), and audio-video (A+V) fine-tuning performance on AudioSet (AS) and VGGSound, alongside recent baselines.
Notably, \ourmodel achieves new state-of-the-art results in audio-only and audio+video classification tasks across these datasets.
On the balanced AS-20K, unbalanced AS-2M, and VGGSound classification tasks, \ourmodel surpasses CAV-MAE~\cite{cavmae} (both models are with ViT-B encoders) in A, V, and A+V tasks by a large margin. This improvement can be attributed to the benefits of reconstructing \emph{aligned} and \emph{contextualized} representations over raw inputs and the enhanced contrastive learning through both \emph{intra}-modal and \emph{inter}-modal contrast. 
Furthermore, \ourmodel outperforms data2vec~\cite{d2v}, highlighting the advantage of utilizing aligned multimodal contexts (Eq.\eqref{eq:loss:context}) over uni-modal contexts as the reconstruction targets.

The fully self-supervised \ourmodel also demonstrates superior performance to supervised audio-video models such as MBT~\cite{nagrani2021attention} in A and A+V classification tasks on AS-20K, AS-2M, and VGGSound.
For video classification on AudioSet, it is worth noting that the fully self-supervised trained video backbone still lags behind MBT which pre-trained with the supervision from ImageNet-21K (11 times larger than ImageNet). This disparity is a consequence of the presence of noise and irrelevant visual context in AudioSet as also discussed in~\cite{audio_ssl_review}. Such bias could make the visual pre-training sub-optimal on AudioSet. We consider resolving this dataset limitation as the future work.

\paragraph{Transfer to Audio/Speech-only Tasks.}
To access the generalizability of the learned audio representations, we further evaluate the AS-pre-trained \ourmodel by transferring it to other out-of-domain speech-only or audio-only tasks.  
Specifically, we conduct experiments on the Environmental Sound Classification (ESC-50)~\cite{piczak2015dataset} and Speech Commands (SPC-v1)~\cite{speechcommandsv2}, where only the audio branch of \ourmodel is fine-tuned.
The results, presented in Table~\ref{tab:audio_task}, demonstrate that \ourmodel outperforms recent supervised and self-supervised models, establishing a new state-of-the-art performance on these benchmarks.
These findings indicate the desirable transferability of \ourmodel from audio-video self-supervised pre-training to audio-only downstream tasks.

\begin{table}[ht]
    \centering\footnotesize
    \parbox{.5\linewidth}{
        \centering
        \footnotesize
        \setlength\tabcolsep{3pt}
        \begin{tabular}{lccc}
        Method & PT & ESC-50 & SPC-1 \\
        \midrule
        AST~\cite{gong21b_interspeech} & IN-SL & 88.7 & 95.5 \\
        SS-AST~\cite{ssast} &  AS-SSL & 88.8 & 96.0 \\
        Aud-MAE~\cite{huang2022masked} & AS-SSL & 94.1 & 96.9 \\ 
        \hline
        \ourmodel &  AS-SSL & 94.4 & 97.3 \\
        \rowcolor{gray!20}\ourmodel &  IN-SSL, AS-SSL & \textbf{94.4} & \textbf{97.4} \\ 
        \end{tabular}
        \vspace{2pt}
        \caption{\textbf{Audio-only tasks} (Acc.$\uparrow$)\label{tab:audio_task}}
    }
    \parbox{.45\linewidth}{
        \centering
        \footnotesize
        \setlength\tabcolsep{3pt}
        \begin{tabular}{lccc}
        Method & PT data & MSR-VTT & YouCook \\
        \midrule
        AVLNet~\cite{rouditchenko2021avlnet}
        & HT100M & 20.1 & 30.7 \\
        TVLT~\cite{tlvt}
        &  HT100M & 22.6 & 31.8 \\
        \hline
        \rowcolor{gray!20}\ourmodel &  AS-2M & 22.8 & 32.2  \\
        \ourmodel &  HT-100M & \textbf{23.8} & \textbf{33.1}\\
        \\
        \end{tabular}
        \vspace{2pt}
        \caption{\textbf{Audio-to-video retrieval} (R@1$\uparrow$\label{tab:retrieval})}
    }
    \vspace{-2em}
\end{table}

\paragraph{Audio-to-Video Retrieval.}
\ourmodel learns aligned audio-video representations that are suitable for textless cross-modal retrieval. To verify this, we further conduct audio-to-video retrieval experiments on MSR-VTT~\cite{xu2016msr} and YouCook~\cite{xu2016msr}.
In these tasks, the audio track of a video serves as the query, and the model performs a search over the testing video collection by computing and ranking the similarity between the query embedding and the video embeddings. We fine-tune \ourmodel using the audio-video pairs in the training sets with Eq.\eqref{eq:2nd_avmae}. We report recall@1 on the testing sets.

Following AVLNet~\cite{rouditchenko2021avlnet} and TVLT~\cite{tlvt}, we explore an alternative setup where we pre-train \ourmodel on HowTo100M~\cite{miech2019howto100m}, a dataset consisting of 1.3 million instructional videos, instead of using AudioSet for pre-training.
It's worth noting that there are notable domain differences between the pre-training datasets (AudioSet or HowTo100M) and the downstream datasets (MSR-VTT or YouCook).
Table~\ref{tab:retrieval} demonstrates that \ourmodel surpasses supervised pre-trained AVLNet and self-supervised pre-trained TVLT, achieving new state-of-the-art performance on these tasks.

\section{Conclusion}
\label{sec:conclusion}
We have presented \ourmodel, a self-supervised audio-video representation learning framework where masked autoencoding meets contrastive learning. 
By leveraging an encoder-fusion-decoder architecture, \ourmodel effectively utilizes complementary information from all modalities for masked autoencoding.
It facilitates efficient contrastive learning in both inter-modal and intra-modal scenarios, even with a high 80\% masking ratio.
Furthermore, \ourmodel highlights a novel pre-training task with self-training to predict homogeneous contextualized audio-video representations.
This approach outperforms conventional uni-modal and multimodal MAEs that predict heterogeneous raw inputs.
As a result, \ourmodel achieves state-of-the-art performance across seven audio-video classification and retrieval tasks, as well as audio-only tasks under the scalable self-supervised learning setup.

\newpage
\bibliographystyle{IEEEtranS}
\bibliography{egbib}

\begin{thebibliography}{10}
\providecommand{\url}[1]{#1}
\csname url@samestyle\endcsname
\providecommand{\newblock}{\relax}
\providecommand{\bibinfo}[2]{#2}
\providecommand{\BIBentrySTDinterwordspacing}{\spaceskip=0pt\relax}
\providecommand{\BIBentryALTinterwordstretchfactor}{4}
\providecommand{\BIBentryALTinterwordspacing}{\spaceskip=\fontdimen2\font plus
\BIBentryALTinterwordstretchfactor\fontdimen3\font minus
  \fontdimen4\font\relax}
\providecommand{\BIBforeignlanguage}[2]{{%
\expandafter\ifx\csname l@#1\endcsname\relax
\typeout{** WARNING: IEEEtran.bst: No hyphenation pattern has been}%
\typeout{** loaded for the language `#1'. Using the pattern for}%
\typeout{** the default language instead.}%
\else
\language=\csname l@#1\endcsname
\fi
#2}}
\providecommand{\BIBdecl}{\relax}
\BIBdecl

\bibitem{bert}
J.~Devlin, M.~Chang, K.~Lee, and K.~Toutanova, ``{BERT:} pre-training of deep
  bidirectional transformers for language understanding,'' in \emph{NAACL-HLT},
  2019.

\bibitem{mae}
K.~He, X.~Chen, S.~Xie, Y.~Li, P.~Doll{\'a}r, and R.~Girshick, ``Masked
  autoencoders are scalable vision learners,'' in \emph{CVPR}, 2022.

\bibitem{feichtenhofer2022masked}
C.~Feichtenhofer, H.~Fan, Y.~Li, and K.~He, ``Masked autoencoders as
  spatiotemporal learners,'' in \emph{NeurIPS}, 2022.

\bibitem{huang2022masked}
P.-Y. Huang, H.~Xu, J.~Li, A.~Baevski, M.~Auli, W.~Galuba, F.~Metze,
  C.~Feichtenhofer \emph{et~al.}, ``Masked autoencoders that listen,'' in
  \emph{NeurIPS}, 2022.

\bibitem{clip}
A.~Radford, J.~W. Kim, C.~Hallacy, A.~Ramesh, G.~Goh, S.~Agarwal, G.~Sastry,
  A.~Askell, P.~Mishkin, J.~Clark, G.~Krueger, and I.~Sutskever, ``Learning
  transferable visual models from natural language supervision,'' in
  \emph{ICML}, 2021.

\bibitem{wei2022contrastive}
Y.~Wei, H.~Hu, Z.~Xie, Z.~Zhang, Y.~Cao, J.~Bao, D.~Chen, and B.~Guo,
  ``Contrastive learning rivals masked image modeling in fine-tuning via
  feature distillation,'' \emph{arXiv preprint arXiv:2205.14141}, 2022.

\bibitem{owens2018audio}
A.~Owens and A.~A. Efros, ``Audio-visual scene analysis with self-supervised
  multisensory features,'' in \emph{ECCV}, 2018.

\bibitem{stica}
\BIBentryALTinterwordspacing
M.~Patrick, P.~Huang, I.~Misra, F.~Metze, A.~Vedaldi, Y.~M. Asano, and J.~F.
  Henriques, ``Space-time crop {\&} attend: Improving cross-modal video
  representation learning,'' in \emph{2021 {IEEE/CVF} International Conference
  on Computer Vision, {ICCV} 2021, Montreal, QC, Canada, October 10-17,
  2021}.\hskip 1em plus 0.5em minus 0.4em\relax {IEEE}, 2021, pp.
  10\,540--10\,552.
\BIBentrySTDinterwordspacing

\bibitem{nce}
\BIBentryALTinterwordspacing
M.~Gutmann and A.~Hyv{\"{a}}rinen, ``Noise-contrastive estimation: {A} new
  estimation principle for unnormalized statistical models,'' in
  \emph{Proceedings of the Thirteenth International Conference on Artificial
  Intelligence and Statistics, {AISTATS} 2010, Chia Laguna Resort, Sardinia,
  Italy, May 13-15, 2010}, ser. {JMLR} Proceedings, vol.~9.\hskip 1em plus
  0.5em minus 0.4em\relax JMLR.org, 2010, pp. 297--304.
\BIBentrySTDinterwordspacing

\bibitem{info_nce}
\BIBentryALTinterwordspacing
A.~van~den Oord, Y.~Li, and O.~Vinyals, ``Representation learning with
  contrastive predictive coding,'' \emph{CoRR}, vol. abs/1807.03748, 2018.
\BIBentrySTDinterwordspacing

\bibitem{beit}
H.~Bao, L.~Dong, and F.~Wei, ``{BEiT}: {BERT} pre-training of image
  transformers,'' in \emph{ICLR}, 2021.

\bibitem{dalle}
\BIBentryALTinterwordspacing
A.~Ramesh, M.~Pavlov, G.~Goh, S.~Gray, C.~Voss, A.~Radford, M.~Chen, and
  I.~Sutskever, ``Zero-shot text-to-image generation,'' in \emph{Proceedings of
  the 38th International Conference on Machine Learning, {ICML} 2021, 18-24
  July 2021, Virtual Event}, ser. Proceedings of Machine Learning Research,
  vol. 139.\hskip 1em plus 0.5em minus 0.4em\relax {PMLR}, 2021, pp.
  8821--8831.
\BIBentrySTDinterwordspacing

\bibitem{maskfeat}
C.~Wei, H.~Fan, S.~Xie, C.-Y. Wu, A.~Yuille, and C.~Feichtenhofer, ``Masked
  feature prediction for self-supervised visual pre-training,'' in \emph{CVPR},
  2022.

\bibitem{Dalal2005}
\BIBentryALTinterwordspacing
N.~Dalal and B.~Triggs, ``Histograms of oriented gradients for human
  detection,'' in \emph{2005 {IEEE} Computer Society Conference on Computer
  Vision and Pattern Recognition {(CVPR} 2005), 20-26 June 2005, San Diego, CA,
  {USA}}.\hskip 1em plus 0.5em minus 0.4em\relax {IEEE} Computer Society, 2005,
  pp. 886--893.
\BIBentrySTDinterwordspacing

\bibitem{dino}
M.~Caron, H.~Touvron, I.~Misra, H.~J\'egou, J.~Mairal, P.~Bojanowski, and
  A.~Joulin, ``Emerging properties in self-supervised vision transformers,'' in
  \emph{ICCV}, 2021.

\bibitem{roy1999learning}
D.~Roy, ``Learning from sights and sounds: a computational model,'' \emph{PhD
  Thesis, MIT Media Laboratory}, 1999.

\bibitem{ramachandram2017deep}
D.~Ramachandram and G.~W. Taylor, ``Deep multimodal learning: A survey on
  recent advances and trends,'' \emph{IEEE signal processing magazine},
  vol.~34, no.~6, pp. 96--108, 2017.

\bibitem{chrupala2022visually}
G.~Chrupa{\l}a, ``Visually grounded models of spoken language: A survey of
  datasets, architectures and evaluation techniques,'' \emph{Journal of
  Artificial Intelligence Research}, vol.~73, pp. 673--707, 2022.

\bibitem{chen1998audio}
T.~Chen and R.~R. Rao, ``Audio-visual integration in multimodal
  communication,'' \emph{Proceedings of the IEEE}, vol.~86, no.~5, pp.
  837--852, 1998.

\bibitem{potamianos2003recent}
G.~Potamianos, C.~Neti, G.~Gravier, A.~Garg, and A.~W. Senior, ``Recent
  advances in the automatic recognition of audiovisual speech,''
  \emph{Proceedings of the IEEE}, vol.~91, no.~9, pp. 1306--1326, 2003.

\bibitem{aleksic2006audio}
P.~S. Aleksic and A.~K. Katsaggelos, ``Audio-visual biometrics,''
  \emph{Proceedings of the IEEE}, vol.~94, no.~11, pp. 2025--2044, 2006.

\bibitem{ngiam2011multimodal}
J.~Ngiam, A.~Khosla, M.~Kim, J.~Nam, H.~Lee, and A.~Y. Ng, ``Multimodal deep
  learning,'' in \emph{ICML}, 2011.

\bibitem{kim2013deep}
Y.~Kim, H.~Lee, and E.~M. Provost, ``Deep learning for robust feature
  generation in audiovisual emotion recognition,'' in \emph{ICASSP}, 2013.

\bibitem{ephrat2018looking}
A.~Ephrat, I.~Mosseri, O.~Lang, T.~Dekel, K.~Wilson, A.~Hassidim, W.~T.
  Freeman, and M.~Rubinstein, ``Looking to listen at the cocktail party: a
  speaker-independent audio-visual model for speech separation,'' \emph{ACM
  Transactions on Graphics (TOG)}, vol.~37, no.~4, pp. 1--11, 2018.

\bibitem{kazakos2019epic}
E.~Kazakos, A.~Nagrani, A.~Zisserman, and D.~Damen, ``Epic-fusion: Audio-visual
  temporal binding for egocentric action recognition,'' in \emph{ICCV}, 2019.

\bibitem{avslowfast}
F.~Xiao, Y.~J. Lee, K.~Grauman, J.~Malik, and C.~Feichtenhofer, ``Audiovisual
  slowfast networks for video recognition,'' \emph{arXiv preprint
  arXiv:2001.08740}, 2020.

\bibitem{nagrani2021attention}
A.~Nagrani, S.~Yang, A.~Arnab, A.~Jansen, C.~Schmid, and C.~Sun, ``Attention
  bottlenecks for multimodal fusion,'' in \emph{NeurIPS}, 2021.

\bibitem{harwath2018jointly}
D.~Harwath, A.~Recasens, D.~Sur{\'\i}s, G.~Chuang, A.~Torralba, and J.~Glass,
  ``Jointly discovering visual objects and spoken words from raw sensory
  input,'' in \emph{ECCV}, 2018.

\bibitem{aytar2016soundnet}
Y.~Aytar, C.~Vondrick, and A.~Torralba, ``Soundnet: Learning sound
  representations from unlabeled video,'' in \emph{NeurIPS}, 2016.

\bibitem{arandjelovic2017look}
R.~Arandjelovic and A.~Zisserman, ``Look, listen and learn,'' in \emph{ICCV},
  2017.

\bibitem{arandjelovic2018objects}
------, ``Objects that sound,'' in \emph{ECCV}, 2018.

\bibitem{korbar2018cooperative}
B.~Korbar, D.~Tran, and L.~Torresani, ``Cooperative learning of audio and video
  models from self-supervised synchronization,'' \emph{Advances in Neural
  Information Processing Systems}, vol.~31, 2018.

\bibitem{shi2022learning}
B.~Shi, W.-N. Hsu, K.~Lakhotia, and A.~Mohamed, ``Learning audio-visual speech
  representation by masked multimodal cluster prediction,'' in \emph{ICLR},
  2022.

\bibitem{ma2020active}
S.~Ma, Z.~Zeng, D.~McDuff, and Y.~Song, ``Active contrastive learning of
  audio-visual video representations,'' in \emph{ICLR}, 2020.

\bibitem{morgado2021audio}
P.~Morgado, N.~Vasconcelos, and I.~Misra, ``Audio-visual instance
  discrimination with cross-modal agreement,'' in \emph{CVPR}, 2021.

\bibitem{patrick2021compositions}
M.~Patrick, Y.~M. Asano, P.~Kuznetsova, R.~Fong, J.~F. Henriques, G.~Zweig, and
  A.~Vedaldi, ``On compositions of transformations in contrastive
  self-supervised learning,'' in \emph{ICCV}, 2021.

\bibitem{wang2021multimodal}
L.~Wang, P.~Luc, A.~Recasens, J.-B. Alayrac, and A.~v.~d. Oord, ``Multimodal
  self-supervised learning of general audio representations,'' \emph{arXiv
  preprint arXiv:2104.12807}, 2021.

\bibitem{zeng2021contrastive}
Z.~Zeng, D.~McDuff, Y.~Song \emph{et~al.}, ``Contrastive learning of global and
  local video representations,'' in \emph{NeurIPS}, 2021.

\bibitem{recasens2021broaden}
A.~Recasens, P.~Luc, J.-B. Alayrac, L.~Wang, F.~Strub, C.~Tallec,
  M.~Malinowski, V.~P{\u{a}}tr{\u{a}}ucean, F.~Altch{\'e}, M.~Valko
  \emph{et~al.}, ``Broaden your views for self-supervised video learning,'' in
  \emph{ICCV}, 2021.

\bibitem{morgado2021robust}
P.~Morgado, I.~Misra, and N.~Vasconcelos, ``Robust audio-visual instance
  discrimination,'' in \emph{CVPR}, 2021.

\bibitem{cavmae}
Y.~Gong, A.~Rouditchenko, A.~H. Liu, D.~Harwath, L.~Karlinsky, H.~Kuehne, and
  J.~Glass, ``Contrastive audio-visual masked autoencoder,'' \emph{arXiv
  preprint arXiv:2210.07839}, 2022.

\bibitem{hinton2015distilling}
G.~Hinton, O.~Vinyals, and J.~Dean, ``Distilling the knowledge in a neural
  network,'' \emph{NeurIPS Deep Learning and Representation Learning Workshop},
  2015.

\bibitem{tian2019contrastive}
Y.~Tian, D.~Krishnan, and P.~Isola, ``Contrastive representation
  distillation,'' in \emph{ICLR}, 2019.

\bibitem{park2019relational}
W.~Park, D.~Kim, Y.~Lu, and M.~Cho, ``Relational knowledge distillation,'' in
  \emph{CVPR}, 2019.

\bibitem{mishra2018apprentice}
A.~Mishra and D.~Marr, ``Apprentice: Using knowledge distillation techniques to
  improve low-precision network accuracy,'' in \emph{ICLR}, 2018.

\bibitem{cho2019efficacy}
J.~H. Cho and B.~Hariharan, ``On the efficacy of knowledge distillation,'' in
  \emph{ICCV}, 2019.

\bibitem{byol}
J.-B. Grill, F.~Strub, F.~Altch{\'e}, C.~Tallec, P.~H. Richemond,
  E.~Buchatskaya, C.~Doersch, B.~A. Pires, Z.~D. Guo, M.~G. Azar, B.~Piot,
  K.~Kavukcuoglu, R.~Munos, and M.~Valko, ``Bootstrap your own latent: A new
  approach to self-supervised learning,'' in \emph{NeruIPS}, 2020.

\bibitem{mvp}
L.~Wei, L.~Xie, W.~Zhou, H.~Li, and Q.~Tian, ``{MVP}: Multimodality-guided
  visual pre-training,'' in \emph{ECCV}, 2022.

\bibitem{moco}
K.~He, H.~Fan, Y.~Wu, S.~Xie, and R.~Girshick, ``Momentum contrast for
  unsupervised visual representation learning,'' in \emph{CVPR}, 2020.

\bibitem{liu2022exploring}
X.~Liu, J.~Zhou, T.~Kong, X.~Lin, and R.~Ji, ``Exploring target representations
  for masked autoencoders,'' \emph{arXiv preprint arXiv:2209.03917}, 2022.

\bibitem{d2v}
A.~Baevski, W.-N. Hsu, Q.~Xu, A.~Babu, J.~Gu, and M.~Auli, ``Data2vec: A
  general framework for self-supervised learning in speech, vision and
  language,'' in \emph{ICML}, 2022.

\bibitem{vaswani2017attention}
A.~Vaswani, N.~Shazeer, N.~Parmar, J.~Uszkoreit, L.~Jones, A.~N. Gomez,
  {\L}.~Kaiser, and I.~Polosukhin, ``Attention is all you need,''
  \emph{NeurIPS}, 2017.

\bibitem{simclr}
T.~Chen, S.~Kornblith, M.~Norouzi, and G.~Hinton, ``A simple framework for
  contrastive learning of visual representations,'' in \emph{ICML}, 2020.

\bibitem{mocov3}
X.~Chen, S.~Xie, and K.~He, ``An empirical study of training self-supervised
  vision transformers,'' in \emph{ICCV}, 2021.

\bibitem{feichtenhofer2016spatiotemporal}
C.~Feichtenhofer, A.~Pinz, and R.~P. Wildes, ``Spatiotemporal residual networks
  for video action recognition,'' in \emph{NIPS}, 2016.

\bibitem{bachmann2022multimae}
R.~Bachmann, D.~Mizrahi, A.~Atanov, and A.~Zamir, ``{MultiMAE}: Multi-modal
  multi-task masked autoencoders,'' \emph{arXiv preprint arXiv:2204.01678},
  2022.

\bibitem{gemmeke2017audio}
J.~F. Gemmeke, D.~P. Ellis, D.~Freedman, A.~Jansen, W.~Lawrence, R.~C. Moore,
  M.~Plakal, and M.~Ritter, ``Audio set: An ontology and human-labeled dataset
  for audio events,'' in \emph{ICASSP}, 2017.

\bibitem{chen2020vggsound}
H.~Chen, W.~Xie, A.~Vedaldi, and A.~Zisserman, ``{VGGSound}: A large-scale
  audio-visual dataset,'' in \emph{ICASSP}, 2020.

\bibitem{xu2016msr}
J.~Xu, T.~Mei, T.~Yao, and Y.~Rui, ``{MSR-VTT}: A large video description
  dataset for bridging video and language,'' in \emph{CVPR}, 2016.

\bibitem{zhou2017towards}
L.~Zhou, C.~Xu, and J.~J. Corso, ``Towards automatic learning of procedures
  from web instructional videos,'' in \emph{AAAI}, 2018.

\bibitem{tagging_right}
\BIBentryALTinterwordspacing
J.~B. Li, S.~Qu, P.~Huang, and F.~Metze, ``{AudioTagging Done Right}: 2nd
  comparison of deep learning methods for environmental sound classification,''
  \emph{CoRR}, vol. abs/2203.13448, 2022.
\BIBentrySTDinterwordspacing

\bibitem{vggish}
\BIBentryALTinterwordspacing
S.~Hershey, S.~Chaudhuri, D.~P.~W. Ellis, J.~F. Gemmeke, A.~Jansen, C.~Moore,
  M.~Plakal, D.~Platt, R.~A. Saurous, B.~Seybold, M.~Slaney, R.~Weiss, and
  K.~Wilson, ``{CNN} architectures for large-scale audio classification,'' in
  \emph{ICASSP}, 2017.
\BIBentrySTDinterwordspacing

\bibitem{hershey2021benefit}
S.~Hershey, D.~P. Ellis, E.~Fonseca, A.~Jansen, C.~Liu, R.~C. Moore, and
  M.~Plakal, ``The benefit of temporally-strong labels in audio event
  classification,'' in \emph{ICASSP}, 2021.

\bibitem{gong21b_interspeech}
Y.~Gong, Y.-A. Chung, and J.~Glass, ``{AST: Audio Spectrogram Transformer},''
  in \emph{Interspeech}, 2021.

\bibitem{Park2019SpecAugmentAS}
D.~S. Park, W.~Chan, Y.~Zhang, C.-C. Chiu, B.~Zoph, E.~D. Cubuk, and Q.~V. Le,
  ``Specaugment: A simple data augmentation method for automatic speech
  recognition,'' \emph{ArXiv}, vol. abs/1904.08779, 2019.

\bibitem{mixup}
H.~Zhang, M.~Cisse, Y.~N. Dauphin, and D.~Lopez-Paz, ``mixup: Beyond empirical
  risk minimization,'' in \emph{ICLR}, 2018.

\bibitem{maskedfeat}
\BIBentryALTinterwordspacing
C.~Wei, H.~Fan, S.~Xie, C.~Wu, A.~L. Yuille, and C.~Feichtenhofer, ``Masked
  feature prediction for self-supervised visual pre-training,'' \emph{CoRR},
  vol. abs/2112.09133, 2021.
\BIBentrySTDinterwordspacing

\bibitem{Kazakos2021SlowFastAuditory}
E.~Kazakos, A.~Nagrani, A.~Zisserman, and D.~Damen, ``Slow-fast auditory
  streams for audio recognition,'' in \emph{IEEE International Conference on
  Acoustics, Speech and Signal Processing (ICASSP)}, 2021, pp. 855--859.

\bibitem{kong2020panns}
Q.~Kong, Y.~Cao, T.~Iqbal, Y.~Wang, W.~Wang, and M.~D. Plumbley, ``{PANNs}:
  Large-scale pretrained audio neural networks for audio pattern recognition,''
  \emph{IEEE/ACM Transactions on Audio, Speech, and Language Processing},
  vol.~28, pp. 2880--2894, 2020.

\bibitem{chen2022hts}
K.~Chen, X.~Du, B.~Zhu, Z.~Ma, T.~Berg-Kirkpatrick, and S.~Dubnov, ``{HTS-AT}:
  A hierarchical token-semantic audio transformer for sound classification and
  detection,'' in \emph{ICASSP}, 2022.

\bibitem{koutini2021efficient}
K.~Koutini, J.~Schl{\"u}ter, H.~Eghbal-zadeh, and G.~Widmer, ``Efficient
  training of audio transformers with patchout,'' \emph{arXiv preprint
  arXiv:2110.05069}, 2021.

\bibitem{ssast}
Y.~Gong, C.-I. Lai, Y.-A. Chung, and J.~R. Glass, ``{SSAST: Self-Supervised
  Audio Spectrogram Transformer},'' in \emph{AAAI}, 2022.

\bibitem{baade2022mae}
A.~Baade, P.~Peng, and D.~Harwath, ``{MAE-AST}: Masked autoencoding audio
  spectrogram transformer,'' in \emph{Interspeech}, 2022.

\bibitem{wang2020makes}
W.~Wang, D.~Tran, and M.~Feiszli, ``What makes training multi-modal
  classification networks hard?'' in \emph{CVPR}, 2020.

\bibitem{jaegle2021perceiver}
A.~Jaegle, F.~Gimeno, A.~Brock, O.~Vinyals, A.~Zisserman, and J.~Carreira,
  ``Perceiver: General perception with iterative attention,'' in \emph{ICML},
  2021.

\bibitem{fayek2020large}
H.~M. Fayek and A.~Kumar, ``Large scale audiovisual learning of sounds with
  weakly labeled data,'' in \emph{IJCAI}, 2020.

\bibitem{audio_ssl_review}
A.~Mohamed, H.-y. Lee, L.~Borgholt, J.~D. Havtorn, J.~Edin, C.~Igel,
  K.~Kirchhoff, S.-W. Li, K.~Livescu, L.~Maaløe, T.~N. Sainath, and
  S.~Watanabe, ``Self-supervised speech representation learning: A review,''
  \emph{IEEE Journal of Selected Topics in Signal Processing}, vol.~16, no.~6,
  pp. 1179--1210, 2022.

\bibitem{piczak2015dataset}
\BIBentryALTinterwordspacing
K.~J. Piczak, ``{ESC}: {Dataset} for {Environmental Sound Classification},'' in
  \emph{Proceedings of the 23rd {Annual ACM Conference} on {Multimedia}}.\hskip
  1em plus 0.5em minus 0.4em\relax {ACM Press}, 2015, pp. 1015--1018.
\BIBentrySTDinterwordspacing

\bibitem{speechcommandsv2}
\BIBentryALTinterwordspacing
P.~{Warden}, ``{Speech Commands: A Dataset for Limited-Vocabulary Speech
  Recognition},'' \emph{ArXiv e-prints}, Apr. 2018.
\BIBentrySTDinterwordspacing

\bibitem{rouditchenko2021avlnet}
A.~Rouditchenko, A.~Boggust, D.~Harwath, B.~Chen, D.~Joshi, S.~Thomas,
  K.~Audhkhasi, H.~Kuehne, R.~Panda, R.~Feris \emph{et~al.}, ``{AVLnet:
  Learning audio-visual language representations from instructional videos},''
  in \emph{Interspeech}, 2021.

\bibitem{tlvt}
Z.~Tang, J.~Cho, Y.~Nie, and M.~Bansal, ``{TVLT: Textless Vision-Language
  Transformer},'' in \emph{NeurIPS}, 2022.

\bibitem{miech2019howto100m}
A.~Miech, D.~Zhukov, J.-B. Alayrac, M.~Tapaswi, I.~Laptev, and J.~Sivic,
  ``{HowTo100M}: Learning a text-video embedding by watching hundred million
  narrated video clips,'' in \emph{ICCV}, 2019.

\bibitem{audiocaps}
C.~D. Kim, B.~Kim, H.~Lee, and G.~Kim, ``Audiocaps: Generating captions for
  audios in the wild,'' in \emph{NAACL-HLT}, 2019.

\bibitem{clotho}
K.~Drossos, S.~Lipping, and T.~Virtanen, ``Clotho: an audio captioning
  dataset,'' in \emph{ICASSP 2020 - 2020 IEEE International Conference on
  Acoustics, Speech and Signal Processing (ICASSP)}, 2020, pp. 736--740.

\bibitem{povey2011kaldi}
D.~Povey, A.~Ghoshal, G.~Boulianne, L.~Burget, O.~Glembek, N.~Goel,
  M.~Hannemann, P.~Motlicek, Y.~Qian, P.~Schwarz \emph{et~al.}, ``The kaldi
  speech recognition toolkit,'' in \emph{IEEE 2011 workshop on automatic speech
  recognition and understanding}, no. CONF.\hskip 1em plus 0.5em minus
  0.4em\relax IEEE Signal Processing Society, 2011.

\bibitem{adamw}
I.~Loshchilov and F.~Hutter, ``Decoupled weight decay regularization,'' in
  \emph{ICLR}, 2019.

\bibitem{sgdr}
------, ``{SGDR}: Stochastic gradient descent with warm restarts,'' in
  \emph{ICLR}, 2017.

\bibitem{droppath}
G.~Larsson, M.~Maire, and G.~Shakhnarovich, ``{FractalNet}: Ultra-deep neural
  networks without residuals,'' in \emph{ICLR}, 2017.

\bibitem{cutmix}
S.~Yun, D.~Han, S.~J. Oh, S.~Chun, J.~Choe, and Y.~Yoo, ``Cutmix:
  Regularization strategy to train strong classifiers with localizable
  features,'' in \emph{ICCV}, 2019.

\bibitem{slowfast}
C.~Feichtenhofer, H.~Fan, J.~Malik, and K.~He, ``Slowfast networks for video
  recognition,'' in \emph{ICCV}, 2019.

\bibitem{mvit}
H.~Fan, B.~Xiong, K.~Mangalam, Y.~Li, Z.~Yan, J.~Malik, and C.~Feichtenhofer,
  ``Multiscale vision transformers,'' in \emph{ICCV}, 2021.

\bibitem{mvitv2}
Y.~Li, C.-Y. Wu, H.~Fan, K.~Mangalam, B.~Xiong, J.~Malik, and C.~Feichtenhofer,
  ``Improved multiscale vision transformers for classification and detection,''
  \emph{arXiv preprint arXiv:2112.01526}, 2021.

\bibitem{clap}
Y.~Wu*, K.~Chen*, T.~Zhang*, Y.~Hui*, T.~Berg-Kirkpatrick, and S.~Dubnov,
  ``Large-scale contrastive language-audio pretraining with feature fusion and
  keyword-to-caption augmentation,'' in \emph{IEEE International Conference on
  Acoustics, Speech and Signal Processing, ICASSP}, 2023.

\bibitem{roberta}
\BIBentryALTinterwordspacing
Y.~Liu, M.~Ott, N.~Goyal, J.~Du, M.~Joshi, D.~Chen, O.~Levy, M.~Lewis,
  L.~Zettlemoyer, and V.~Stoyanov, ``Roberta: {A} robustly optimized {BERT}
  pretraining approach,'' \emph{CoRR}, vol. abs/1907.11692, 2019.
\BIBentrySTDinterwordspacing

\bibitem{Oncescu21a}
\BIBentryALTinterwordspacing
A.~Oncescu, A.~S. Koepke, J.~F. Henriques, Z.~Akata, and S.~Albanie, ``Audio
  retrieval with natural language queries,'' in \emph{Interspeech 2021, 22nd
  Annual Conference of the International Speech Communication Association,
  Brno, Czechia, 30 August - 3 September 2021}.\hskip 1em plus 0.5em minus
  0.4em\relax {ISCA}, 2021, pp. 2411--2415.
\BIBentrySTDinterwordspacing

\bibitem{Mei2021ACT}
X.~Mei, X.~Liu, Q.~Huang, M.~D. Plumbley, and W.~Wang, ``Audio captioning
  transformer,'' in \emph{Proceedings of the 6th Detection and Classification
  of Acoustic Scenes and Events 2021 Workshop (DCASE2021)}, Barcelona, Spain,
  November 2021, pp. 211--215.

\bibitem{gpt}
\BIBentryALTinterwordspacing
T.~B. Brown, B.~Mann, N.~Ryder, M.~Subbiah, J.~Kaplan, P.~Dhariwal,
  A.~Neelakantan, P.~Shyam, G.~Sastry, A.~Askell, S.~Agarwal,
  A.~Herbert{-}Voss, G.~Krueger, T.~Henighan, R.~Child, A.~Ramesh, D.~M.
  Ziegler, J.~Wu, C.~Winter, C.~Hesse, M.~Chen, E.~Sigler, M.~Litwin, S.~Gray,
  B.~Chess, J.~Clark, C.~Berner, S.~McCandlish, A.~Radford, I.~Sutskever, and
  D.~Amodei, ``Language models are few-shot learners,'' in \emph{NeurIPS},
  2020.
\BIBentrySTDinterwordspacing

\bibitem{SahidullahKH15}
\BIBentryALTinterwordspacing
M.~Sahidullah, T.~Kinnunen, and C.~Hanil{\c{c}}i, ``A comparison of features
  for synthetic speech detection,'' in \emph{{INTERSPEECH} 2015, 16th Annual
  Conference of the International Speech Communication Association, Dresden,
  Germany, September 6-10, 2015}.\hskip 1em plus 0.5em minus 0.4em\relax
  {ISCA}, 2015, pp. 2087--2091.
\BIBentrySTDinterwordspacing

\bibitem{chintha2020recurrent}
A.~Chintha, B.~Thai, S.~J. Sohrawardi, K.~Bhatt, A.~Hickerson, M.~Wright, and
  R.~Ptucha, ``Recurrent convolutional structures for audio spoof and video
  deepfake detection,'' \emph{IEEE Journal of Selected Topics in Signal
  Processing}, vol.~14, no.~5, pp. 1024--1037, 2020.

\end{thebibliography}

\newpage

\appendix
\section*{Appendix}
This appendix are organized as follows: In \S\ref{sec:app:vis}, we present the qualitative results of audio and video reconstruction. These results are obtained using the stage-1 \ourmodel's decoders, which are trained to reconstruct raw inputs. In \S\ref{sec:app:hyper}, we offer the comprehensive experimental details and hyperparameter configurations for pre-training and fine-tuning on each dataset. In \S\ref{sec:app:exp}, we perform additional experiments to evaluate and analyze \ourmodel's performance. These experiments include: 
\begin{enumerate}
\item Modality-wise masking ratio and masking type analysis.
\item Contrastive weights/ hyper-parameters analysis.
\item From-scratch and large model analysis.
\item Text-audio retrieval tasks on AudioCaps~\cite{audiocaps} and Clotho~\cite{clotho}.  
\end{enumerate}

In \S\ref{sec:app:limit}, we discuss \ourmodel's societal impact and limitations.


\begin{figure*}[h!]
    \centering
    \includegraphics[width=1.0\linewidth]{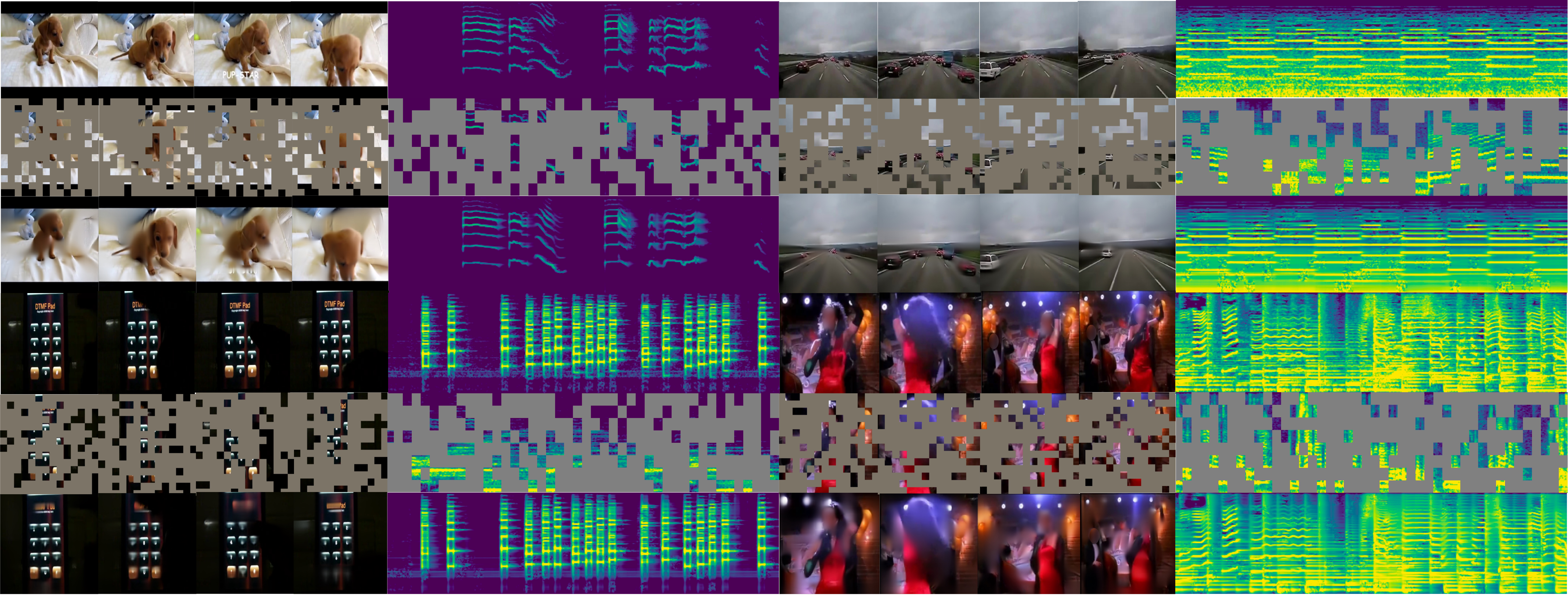} 
    \caption{
        \textbf{Video clip and spectrogram reconstruction on the AudioSet \textit{eval} set}.
        We sample 4 paired (video, audio) examples as follows:
        Top left: a puppy video; 
        Top right: a recording from an ambulance's dash camera;
        Bottom left: a person dialing a phone in a dark room;
        Bottom right: a singer dancing.
        Input masking ratio: 70\%.
        In each 3-row group, we show the original video and its audio spectrogram (top), masked input to \ourmodel (middle), and \ourmodel's video and audio spectrogram reconstructions (bottom). 
        The spectrogram shape is 1024$\times$128; patch size is 16$\times$16. Each spectrogram has 64$\times$8=512 patches. After applying 70\% masking, there are 154 patches visible to \ourmodel. 
        The 8-frame (4-second under 2 fps) video clip size is $8\times3\times224\times224$; patch size is $16\times16$. Each video has $4\times14\times14=784$ patches after patch embedding (temporal kernel/stride=2). After applying 70\% masking, there are 235 patches visible to \ourmodel. 
    }
    \label{fig:vis}
\end{figure*}

\section{Raw Audio-Video Reconstructions}
\label{sec:app:vis}
In Fig. \ref{fig:vis}, we employ a stage-1 \ourmodel (ViT-B) to reconstruct raw audio spectrograms and video frames with masked inputs.
The model is trained using an 80\% masking ratio on the AudioSet-2M full training set with \emph{un-normalized} raw spectrograms and video frames  as the reconstruction targets (Eq.\eqref{eq:1st_avmae}, stage-1). We visualize the reconstruction results by \ourmodel's audio and video decoders, wherein 70\% of the input tokens are masked to its encoders. This visualization is performed on the AudioSet \emph{eval} set.

The results demonstrate that \ourmodel effectively reconstructs highly corrupted versions of both audio spectrograms and video frames in video clips. The generated reconstructions for videos exhibit high fidelity and preserve spatial and temporal consistency of visual objects (\eg, the nearby moving cars recorded by the ambulance’s dash camera) across different input domains, scenes, and lighting conditions.
In the case of audio reconstructions, \ourmodel accurately maintains the positions and arrangements of time-frequency components in the spectrogram (\eg, the ambulance’s siren and the song by the singer), which are essential for human understanding and perception of sound. Furthermore, the reconstructed audio and video components are consistent and well-aligned in time, enhancing the overall coherence of the reconstructed content.

\section{Experimental Details \& Hyper-parameters}
\label{sec:app:hyper}

In this section, we provide additional experimental details for data preprocessing, implementation, pre-training, fine-tuning, and inference. The hyper-parameters are summarized in Table~\ref{tab:app:hyper}. The codebase and the pre-trained models will be available.

\subsection{Data Preprocessing}
In our study, we obtained a total of 2.01 million AudioSet videos, including both the video and audio tracks from the balanced and unbalanced training set and the evaluation set. Additionally, we managed to collect 198K VGGSound videos.
As part of the preprocessing, we resized the video tracks to 360p while maintaining the aspect ratio and adjusting the longer dimension to 360 pixels. We also resampled the audio tracks to a sampling rate of 16K.
We employed different temporal footprints for modeling the audio and video in \ourmodel, specified as the following:

Following the preprocessing in \cite{nagrani2021attention,gong21b_interspeech,huang2022masked}, we transform a raw audio (with mono-channel and under 16K sampling rate) into 128 Mel-frequency bands used in Kaldi~\cite{povey2011kaldi}. This transformation involves using a 25ms Hanning window that shifted every 10ms. We then normalize the spectrogram according to the mean and variance in each dataset.
For a 10-second audio, the resulting spectrogram has a dimension of $1024\times128$.

Regarding the video part, we utilize 4-second clips consisting of 8 frames captured at a rate of 2 frames per second (fps). Each input frame has a size of $224\times224$. In the pre-training phase, we apply common data augmentations such as random horizontal flip (with a probability of 0.5) and multi-scale random cropping (with a scale ranging from 0.2 to 1.0). In contrast, we apply only center cropping during the testing or inference phase.
When processing a 10-second video clip from AudioSet, we randomly sample a starting point and extracted the consecutive 4 seconds of the video (cyclically looping back to the beginning if it was shorter than 4 seconds). As a result, the video clip input, consisting of 3 channels, had dimensions of $8\times3\times224\times224$.

\subsection{Implementation}
\paragraph{Uni-modal Encoders.}
We adopt the main design choices from original MAE for images \cite{mae} and Audio-MAE \cite{huang2022masked}. Specifically, we employ separate 12-layer Transformers with 12 attention heads as the encoders for each modality. The patch embedding and positional embeddings layers are also separated for each modality.
During our investigation, we explored alternative designs, including sharing the audio-video encoder weights with separated inputs or concatenating them as done in Multi-MAE \cite{bachmann2022multimae}. However, these alternative architectures resulted in inferior performance compared to the proposed architecture of using separated encoders with separated inputs. As a result, we chose to adhere to the original design of separate encoders for each modality.

In all Transformer encoders (with ViT-B as the default), the embedding dimension $H$ is set to 768
For each input spectrogram of size $1024\times128$ representing a 10-second audio, we tokenize it into non-overlapping $16\times16$ spectrogram patches using an audio patch embedding layer.
The kernel and stride sizes for both the time and frequency dimensions are 16, resulting in a total of $64\times8$ spectrogram patches or tokens for the audio sequence.
The flattened audio token sequence has a length $N$ of 512. Each audio token corresponds to a 768-dimensional vector. After appending the \texttt{[CLS]} token, adding positional embeddings, and applying 80\% masking, the final input audio token sequence is represented as $\mathbf{a'}\in\mathbb{R}^{102\times768}$.

For each video clip with dimensions $8\times3\times224\times224$ (4 seconds in duration), we tokenize it into non-overlapping cells using a video patch embedding layer. 
The spatial kernel and stride sizes are set to 16, while the temporal kernel and stride sizes are set to 2. This process results in a total of $4\times14\times14=784$ video patches or tokens.
The flattened video token sequence has a length $M$ of 784. Each video token corresponds to a 768-dimensional vector. After appending the \texttt{[CLS]} token, adding positional embeddings, and applying 80\% masking, the final input video token sequence is represented as $\mathbf{v'}\in\mathbb{R}^{156\times768}$.

\paragraph{Fusion Encoders.}
Following the ViT-B uni-modal encoders, we incorporate an audio-video \textit{fusion} encoder. The fusion encoder consists of a two-layer (with $L$=2) Transformer, which can be either a vanilla Transformer or an MBT Transformer \cite{nagrani2021attention}.

In the vanilla Transformer setup, the fusion encoder, denoted as $g_{\text{av}}(\cdot)$, jointly encodes the audio and video tokens. This is done by concatenating the output of the uni-modal encoders for audio ($\mathbf{a}{\text{um}}^{l+1}$) and video ($\mathbf{v}{\text{um}}^{l+1}$) as input, resulting in $(\mathbf{a}_{\text{um}}^{l+1}\|\mathbf{v}_{\text{um}}^{l+1})=\text{Transformer}^{l}(\mathbf{a}_{\text{um}}^l\|\mathbf{v_{\text{um}}}^l)$, where ``$\|$'' denotes concatenation.

In the MBT setup, we extend the vanilla Transformer by appending an additional 4 trainable MBT tokens for each modality. MBT encourages the model to more selectively collate and condense relevant information in each modality by forcing information exchange between modalities to pass through a small number of learnable bottleneck features $\mathbf{b}^0=[b_1 \dots b_4], b_i \in \mathbb{R}^H$. The use of MBT tokens was originally proposed in the context of supervised audio-video learning.
Precisely, $\mathbf{a}_{\text{um}}^{l+1}\| \mathbf{b}_{\text{a}}^{l+1}=g_{\text{av}}^l(\mathbf{a}_{\text{um}}^l\|\mathbf{b}^l)$ 
and $\mathbf{v}_{\text{um}}^{l+1}\| \mathbf{b}_{\text{v}}^{l+1}=g_{\text{av}}^{l}(\mathbf{v}_{\text{um}}^l\|\mathbf{b}^l)$, 
where $\mathbf{b}^{l+1}=(\mathbf{b}_{\text{a}}^{l+1}+\mathbf{b}_{\text{v}}^{l+1})/2$. 

\paragraph{Decoders.}
The audio and video decoders are 8-layer Transformers with an embedding dimension of 512 and 16 attention heads. In the top decoder layer, we applied a linear prediction head to either predict the raw audio spectrogram and video frame patches in stage-1 (\ie, $\mathbf{a}^{\text{raw}} \in \mathbb{R}^{H_{\text{raw}}^{\text{a}}}$ and 
$\mathbf{v}^{\text{raw}} \in \mathbb{R}^{H_{\text{raw}}^{\text{v}}}$), or predict the aliened and contextualized representations in stage-2 (\ie $\mathbf{a}^{\text{Teacher}}, \mathbf{v}^{\text{Teacher}}, \mathbf{\tilde{a}}, \mathbf{\tilde{v}} \in\mathbb{R}^{H}$).
The audio/video encoder and decoder in \ourmodel have 86M and 27M parameters, respectively.
The floating point operations (FLOPs) for the audio encoder are 48.6G, comparable to the audio encoders in Audio-MAE~\cite{huang2022masked} and CAV-MAE~\cite{cavmae}.

\begin{table*}[t]\centering%
    \tablestyle{2pt}{1.1}
    \setlength\tabcolsep{4.0pt}
    \begin{tabular}{l|c|ccccc}
        & Pre-training & \multicolumn{5}{c}{Fine-tuning} \\
        Configuration & AS-2M PT  & AS-2M  & AS-20K  & VGGSound & ESC & SPC\\
        \toprule
        Optimizer & \multicolumn{6}{c}{AdamW~\cite{adamw}}\\
        Optimizer momentum & \multicolumn{6}{c}{$\beta_1=0.9$, $\beta_2=0.95$}\\
        Weight decay & \multicolumn{6}{c}{0.00001} \\
        Base learning rate & 0.0002 & 0.0001\textsuperscript{$\dagger$} & 0.001 & 0.0002 & 0.0005 & 0.001  \\
        Learning rate schedule & \multicolumn{6}{c}{half-cycle cosine decay~\cite{sgdr}}\\
        Minimum learning rate & \multicolumn{6}{c}{0.000001}\\
        Gradient clipping & \multicolumn{6}{c}{None}\\
        Warm-up epochs & 4 & 20  & 4  & 4 & 4 & 1 \\
        Epochs & 20 & 100 & 60 & 60 & 60 & 10  \\
        Batch size & 512 & 512 & 64 & 256 & 64 & 256 \\
        GPUs & 64 & 64 & 8 & 32 & 4 & 4 \\
        Weighted sampling  & False & True & False & True & False &  False\textsuperscript{*} \\
        Weighted sampling  size & - & 200,000 & - & 200,000 & -& - \\
        Augmentation & R & R & R & R+N & R & R+N   \\
        SpecAug~\cite{Park2019SpecAugmentAS} (time/frequency) & - & 192/48 & 192/48 & 192/48 &  96/24 & 48/48  \\
        Drop path~\cite{droppath} & 0.0 & 0.1 & 0.1 & 0.1 & 0.1 & 0.1   \\
        Mixup~\cite{mixup} & 0.0 & 0.5 & 0.5 & 0.5 & 0.0 & 0.5   \\
        Multilabel &n/a & True & True & False & False & False  \\
        Loss Function & MSE & BCE & BCE & BCE & CE & BCE   \\
        Dataset Mean for Normalization & -4.268 & -4.268 & -4.268 & -5.189 & -6.627  & -6.702  \\
        Dataset Std for Normalization & 4.569 & 4.569 & 4.569 & 3.260 & 5.359  & 5.448 \\
    \end{tabular}
    \caption{\textbf{Pre-training (PT) and Fine-tuning (FT) hyper-parameters}. For augmentation, R: sampling random starting points with cyclic rolling in time; N: adding random noise (signal-to-noise ratio (SNR): 20dB) to spectrograms. For loss functions, BCE: binary cross entropy loss (for multi-label datasets or when using mixup); CE: cross-entropy loss, MSE: mean square error loss.
    \textsuperscript{*}: We repeat and balance each class to 50\% of the size of the unknown class.    
    \textsuperscript{$\dagger$}: For ViT-S, We use a learning rate of 0.0005 on AS-2M FT and 0.002 on AS-20K FT for the ViT-S model. For the ViT-L model, we use 0.0001 and 0.0005 for AS-2M and AS-20K FT experiments.
    \label{tab:app:hyper}}
\end{table*}

\subsection{Training and Inference}

\paragraph{Pre-training.}
\ourmodel operates under a fully self-supervised learning setup for pre-training. 
For pre-training \ourmodel's audio branch, we randomly initialize it from scratch.
For the visual branch, we either randomly initialize it or initialize it with the self-supervised MAE~\cite{mae} pre-trained on ImageNet where we simply repeat and inflate the convolution kernel in its patch-embedding to handle the additional temporal domain.
Different visual initialization methods are compared in Table~\ref{tab:as_result} in the main paper and Table~\ref{app:tab:init} in Appendix.
Importantly, \ourmodel operates under the fully \emph{self-supervised} setup.

\ourmodel is pre-trained on the combined unbalanced and balanced training sets of AS-2M. The pre-training process is performed using 64 GPUs with a 512 accumulated batch size. 
In stage-1 and each iteration of stage-2 (for $K=3$ iterations), we pre-train the model for 20 epochs. Each pre-training session takes approximately 20 hours to complete. In total, the pre-training process takes around 80 hours.
Note that the effective learning rate ($lr_{\text{eff}}$) depends on the base learning rate ($lr_{\text{base}}$) and the batch size. Precisely, $lr_{\text{eff}}=lr_{\text{base}}*\frac{\text{batch size}}{256}$.
In our experiments, we also tried using strong data augmentations (\eg, mixup~\cite{mixup}, SpecAug~\cite{mixup}, and CutMix~\cite{cutmix}) to augment audio spectrograms during the pre-training phase. However, we observed that the resulting performance was either similar or worse compared to the baseline. Therefore, by default, we exclude these strong data augmentations for both audio and video during the pre-training phase.


\paragraph{Fine-tuning.}
We fine-tune \ourmodel in three scenarios: (1) audio-only, (2) video-only, and (3) audio+video.
We follow the setup in MAE and retain only the pre-trained uni-modal encoders for fine-tuning.
In the audio-only and video-only setups, we fine-tune the respective encoders in the \ourmodel (stage-2). 
In the audio+video fusion setup, we introduce a 2-layer vanilla Transformer on top of the audio and video encoder in the \ourmodel (stage-2) and fine-tune it using both audio and video inputs.
The hyperparameter configurations specified in Table~\ref{tab:app:hyper} are employed for finetuning on each dataset.
Empirically we observed a discrepancy in convergence rate between audio and video. We circumvent this by applying a 50\% learning rate reduction for the weights of the video encoder when performing audio+video fusion fine-tuning.  

We adopt the standard fine-tuning pipeline and augmentation in prior audio/audio-video classification works~\cite{gong21b_interspeech, huang2022masked, nagrani2021attention}. 
Specifically, we employ SpecAug~\cite{Park2019SpecAugmentAS}, mixup~\cite{mixup}, balanced sampling~\cite{tagging_right}, and fine-tuning masking~\cite{huang2022masked} (a 20\% random masking rate for time and frequency in audio spectrograms; 20\% for space and time in video clips). 
For video, we use standard video augmentations used in video classification~\cite{maskedfeat, feichtenhofer2022masked}.

To perform importance sampling that balance the fine-tuning scheme on the unbalanced AS-2M (and VGGSound),
we apply a distributed weighted sampler as prior works~\cite{tagging_right, gong21b_interspeech, chen2022hts, koutini2021efficient}. 
We set the probability of sampling a sample proportional to the inverse frequency of its labels, where the label frequency is estimated over the training set. Specifically, for a instance $i$ in a dataset $\mathcal{D}$ with a label pool $\mathbf{C}$, its sampling weight is proportional to $\sum_{c_i \in \mathbf{C}}{w_c}$, where $w_c=\frac{1000}{\sum_{i\in\mathbf{D}}{c_i}+\epsilon}$ and $\epsilon=0.01$ is set to avoid underflow in majority classes.
During the fine-tuning process on AS-2M, we randomly sample 200K instances (approximately 10\% of AS-2M) with replacement in each epoch. We fine-tune \ourmodel for 100 epochs, which corresponds to approximately 10 full epochs of AS-2M. The entire fine-tuning process typically takes around 10 hours to complete.

\paragraph{Inference.}
After fine-tuning, we select the last checkpoint for inference. For the video and audio+video tasks, we adopt the standard approach used in video action recognition~\cite{slowfast, mvit, mvitv2} by uniformly sampling ten 4-second video clips throughout the time domain of a video. Each of these sampled video clips is individually fed forward through the model to generate predictions. Note that for audio+video classification, the audio input remains the same 10-second audio recording throughout the sampling of video clips.

\begin{table}[!h]
\centering
\setlength\tabcolsep{3.0pt}
\footnotesize
\begin{tabular}{c|x{35}>{\columncolor{defaultcolor}}x{35}c}
\# Clips (AS-2M) 
& 1 & 10 \\
\shline
Audio & 48.7 & 48.7 \\
Video & 29.4 & 30.3 \\
Audio+Video & 52.6 & 53.3 \\
\end{tabular}
\caption{\textbf{Number of video clips in the inference time.}\label{tab:clip_aggregation}}
\end{table}

We average the ten predictions as the instance-level prediction and report the classification performance in Table~\ref{tab:as_result} in \S\ref{sec:exp}. Note that these results are based on single-modal predictions, without ensembling multiple models. In Table~\ref{tab:clip_aggregation}, we compare the results obtained from one-clip predictions and ten-clip predictions (mAP on AS-2M). The sampling of ten clips leads to improvements of up to 0.9 mAP for video-only and audio+video tasks, while the audio-only task remains unaffected.


\section{Additional Experiments and Analysis}
\label{sec:app:exp}
In this section, we present additional analysis to extend the study of the module-wise contribution in Table~\ref{app:tab:ablation}. We then expand our study on another important type of audio task: text-audio retrieval.

We organize this section as follows: Firstly, we investigate how different choices of masking ratio and masking type may affect the model performance.
Next, we examine the effects of adjusting contrastive weights in the training objective. By exploring different weight settings, we aim to understand the influence of contrastive learning on the model's ability to capture audio-video relationships.
Furthermore, we compare different approaches to visual backbone initialization and evaluate the performance using larger (ViT-L) audio/video encoders in \ourmodel-Large models. This analysis helps us understand the benefits and trade-offs of using larger backbone models and different initialization strategies.
Additionally, besides audio-video classification tasks and audio-video retrieval tasks presented in the main paper. We include our study on audio-text retrieval tasks in the last.

\begin{table}[!h]
\centering
\setlength\tabcolsep{2.0pt}
\footnotesize
\begin{tabular}{lll}
Method & Audio & Video \\
\shline
A-MAE/V-MAE (baseline) & 36.4 & 17.4 \\ \hline
\multicolumn{3}{l}{\textit{\ourmodel stage-1}}\\
+ Joint AV-MAE & 36.8\textsubscript{(\tiny+0.4)} & 17.7\textsubscript{(\tiny+0.3)} \\ 
+ Intra and Inter contrast & 39.0\textsubscript{(\tiny+2.2)} & 22.2\textsubscript{(\tiny+4.5)} \\ \hline
\multicolumn{3}{l}{\textit{\ourmodel stage-2}}\\
\rowcolor{gray!20} + Student-teacher learning & 41.8\textsubscript{(\tiny+2.8)} & 24.8\textsubscript{(\tiny+2.6)} \\        
\end{tabular}
\vspace{3pt}
\caption{\textbf{Module-wise Contribution} in \ourmodel).\label{app:tab:ablation}}
\end{table}

\subsection{Masking Ratio and Type}
In addition to applying a shared masking ratio for each modality, we also investigated the impact of applying different masking ratios for audio and video. The results of this analysis are summarized in Table~\ref{app:tab:masking_ratio}. Interestingly, we did not observe a significant change in performance (mAP on AS-20K) when using different masking ratios for audio and video. Based on these findings, we simplify the approach by defaulting to an 80\% masking ratio for both audio and video, as the Joint AV-MAE entry (the second row) in Table~\ref{app:tab:ablation}.

\begin{table}[!h]
    \centering
    \subfloat[\textbf{Modality-wise Masking}\label{app:tab:masking_ratio}]{       
        \makebox[0.35\linewidth][c]{
            \setlength\tabcolsep{3.0pt}
            \tablestyle{2pt}{1.05}
            \footnotesize
            \begin{tabular}{cccc}
            Ratio & 70\% (A) & 80\% (A) & 90\% (A) \\
            \hline
            70\% (V) & 36.7/17.5 & 36.8/17.5 &  36.4/17.3 \\
            80\% (V) &  36.7/17.2 & 36.8/17.7 & 36.8/17.4  \\
            90\% (V) &  36.5/17.3 & 36.6/17.6 & 36.8/17.5  \\
            \\
            
            \end{tabular}  
        }
    }
    \subfloat[\textbf{Masking Type}\label{app:tab:masking_type}]{       
        \makebox[0.65\linewidth][c]{
            \setlength\tabcolsep{3.0pt}
            \tablestyle{2pt}{1.05}
            \footnotesize
            \begin{tabular}{cccc}
            Type & 70\% & 80\% & 90\% \\
            \hline
            Random (A), Random (V) &  36.7/17.5 & 36.8/17.7 & 36.8/17.5  \\
            Time-Freq (A), Random (V) &  36.2/17.5 & 36.3/17.7 & 36.3/17.8  \\
            Random (A), Space-Time (V) &  36.7/17.2 & 36.7/17.3 & 36.8/17.5  \\
            Time-Freq (A), Space-Time (V) &  36.0/17.1 & 36.2/17.1 & 36.3/17.3  \\
            \end{tabular}  
        }
    }
\caption{\textbf{Masking Ratio} and \textbf{Masking Type} (mAP on AS-20K).\label{app:tab:masking}}
\end{table}

The default masking strategy in our model is random masking, which applies the same Bernoulli trial parameterized by a masking ratio ($p$) to each spectrogram or RGB patch. In Table~\ref{app:tab:masking_type}, we explored more advanced masking strategies and compare their impacts.
For audio spectrogram, in addition to random masking (time-and-frequency agnostic with Bernoulli trials), we investigated time-masking (randomly masks multiple periods of time components) and frequency masking (randomly masks multiple frequency bands). We perform Bernoulli trials on time or frequency slots instead of individual patches. 
For video frames, we explored time-wise masking (randomly masking an entire frame) and space-wise masking (randomly masking a spatial patch across time). We set the masking ratio between spatial/frequency and time as 2:1 and adjusted the overall ratio from 70\% to 90\% for comparison with random masking.

Surprisingly, we do not observe improvements when applying these advanced masking strategies for multimodal pre-training. The simplest random masking approach achieved the best pre-training performance. This observation aligns with the findings in uni-modal MAEs~\cite{mae,feichtenhofer2022masked,huang2022masked}, suggesting that the random masking strategy is effective and sufficient for multimodal pre-training.

\subsection{Contrastive Weights}
Table~\ref{app:tab:contrastive} showcases the impact of adjusting contrastive weights $\alpha$ and $\beta$ in \ourmodel. The results show that fine-tuning these contrastive weights leads to improved performance. In our experiments, we set $\alpha=0.1$ and $\beta=0.01$ which yield the best performance.

It is important to note that the smaller contrastive weights in Eq.\eqref{eq:1st_avmae} do not imply that the contrastive objectives are less significant. The weights are chosen to scale and balance the gradients from the reconstruction and the two contrastive objectives to ensure they fall within a comparable range. This adjustment enhances training stability.
Furthermore, the softmax temperatures used in NCE (Eq.~\eqref{equ:contrastive}) are set as $\tau_{\text{c}}^{\text{inter}}=0.1$ (more tolerant) for inter-modal contrastive learning and $\tau_{\text{c}}^{\text{intra}}=1.0$ (stricter) for intra-modal contrastive learning. These temperature values help regulate convergence across modalities in the contrastive learning process.

\begin{table}[!h]
    \centering
    \subfloat[\textbf{Inter-modal $\alpha$}~\label{app:tab:contrastive:alpah}]{       
        \makebox[0.3\linewidth][c]{
            \setlength\tabcolsep{2.0pt}
            \tablestyle{2pt}{1.05}
            \footnotesize
            \begin{tabular}{c|x{20}>{\columncolor{defaultcolor}}x{20}x{20}}
            $\alpha$ & 0.3 & 0.1 &  0.05 \\
            \shline
            Audio & 41.5 & 41.8 & 41.4 \\
            Video & 24.3 & 24.8 & 24.4 \\
            \end{tabular}     
        }
    }
    \subfloat[\textbf{Intra-modal $\beta$}~\label{app:tab:contrastive:beta}]{       
        \makebox[0.3\linewidth][c]{
            \setlength\tabcolsep{2.0pt}
            \tablestyle{2pt}{1.05}
            \footnotesize
            \begin{tabular}{c|x{20}x{20}>{\columncolor{defaultcolor}}x{20}}
            $\beta$ & 0.1 & 0.05 &  0.01 \\
            \shline
            Audio & 41.3 & 41.5 & 41.8 \\
            Video & 24.3 & 24.7 & 24.8 \\
            \end{tabular}     
        }
    }
\caption{\textbf{Contrastive Weights} (mAP on AS-20K).\label{app:tab:contrastive}}
\end{table}

\subsection{From-scratch Visual Backbone and Large Models}
Under the fully self-supervised setup, \ourmodel initializes its audio branch from scratch and initialize its visual branch either from scratch or from a ImageNet self-supervised pre-trained MAE (IN-SSL).
In this part, we further explore and compare the visual backbone initialization strategies under different model sizes.

\begin{table}[t!]
\setlength\tabcolsep{4.0pt}
\small
\centering
\footnotesize
\begin{tabular}{@{}lcccccccc@{}}
 & & & \multicolumn{3}{c}{AS-20K} & \multicolumn{3}{c}{AS-2M}\\ \cmidrule(l){4-9} 
Model & A-init & V-init & A & V & A+V & A & V & A+V \\ \midrule
\rowcolor{gray!20} \ourmodel-Base & scratch & IN-SSL & 41.8 & 24.8 & 
44.9 & 48.7 & 30.3 & 53.3 \\
\ourmodel-Base & scratch & scratch & 41.6 & 23.7 & 44.6 & 48.7 & 28.3 & 51.9 \\
\ourmodel-Large & scratch & IN-SSL & 42.1 & 27.1 & 45.3 & 48.8 & 32.4 & 53.3 \\
\ourmodel-Large & scratch & scratch & 42.3 & 25.3 & 45.1 & 49.1 & 30.6 & 52.5 \\
\end{tabular}
\vspace{3pt}
\caption{
\textbf{Visual Backbone Initialization and Model Size} (mAP).\label{app:tab:init}}
\end{table}

As shown in the top two rows of Table~\ref{app:tab:init}, when considering \ourmodel-Base models, there is a small gap (-0.2 mAP on AS-20K) observed in the audio stream when discarding visual initialization from the ImageNet self-supervised model.
However, a larger gap (-0.9 mAP) is observed in the video stream. 
A similar trend is observed in the AS-2M experiments. This discrepancy in the visual part can likely be attributed to biases and visual quality issues such as misalignment, title-only content, and low-resolution videos present in AudioSet.

To address this gap in the visual part, incorporating additional uni-modal pre-training steps could potentially improve model performance. For instance, conducting separate audio-only and video-only large-scale pre-training as the first step.
In this work, we focus on audio-video pre-training solely on AudioSet for simplicity and for fair comparison with baselines. The possibility of incorporating additional pre-training steps is left for future research.

When using large models (ViT-L, rows 3-4), the gap in visual mAP (-1.8 mAP) still persists. Interestingly, the audio part of large models actually benefits from from-scratch visual initialization, showing an improvement of +0.2-0.3 mAP. Additionally, when comparing rows 1-2 to rows 2-3, the visual stream is benefited more by employing a larger (ViT-L) backbone.
Across all the configurations (from-scratch or visual initialization with IN-SSL), \ourmodel consistently outperforms recent baselines (in Table~\ref{tab:as_result} of the main paper) by a significant margin.

\subsection{Text-Audio Tasks}
Another important audio-centered multimodal application involves text-to-audio and audio-to-text retrieval tasks.
In text-to-audio retrieval, the query is a text description, and the model performs a search over the (testing) audio collection by computing and ranking the similarity between the query embedding and the audio embeddings.
To evaluate the audio representations learned by \ourmodel, following CLAP~\cite{clap}, we add a text encoder initialized from Roberta~\cite{roberta}. We perform fine-tuning with inter-modal contrast on the same training set used by CLAP. Specifically, AudioCaps~\cite{audiocaps} and Clotho~\cite{clotho}, and LAION-630K~\cite{clap}. In Table~\ref{tab:clap}, we report recall@1, 5, and 10 on the testing sets.

\begin{table}[h!]
\setlength\tabcolsep{3.0pt}
\small
\centering
\footnotesize
\begin{tabular}{lcccccccccccc}
  & \multicolumn{6}{c}{AudioCaps~\cite{audiocaps}} & \multicolumn{6}{c}{Clotho~\cite{clotho}}\\ \cmidrule(l){2-13}
& \multicolumn{3}{c}{Text-to-Audio} & \multicolumn{3}{c}{Audio-to-Text} & \multicolumn{3}{c}{Text-to-Audio} & \multicolumn{3}{c}{Audio-to-Text} \\ \cmidrule(l){2-13}
Model &  R@1 & R@5 & R@10 & R@1 & R@5 & R@10 & R@1 & R@5 & R@10 & R@1 & R@5 & R@10 \\ \midrule
MMT\textsuperscript{*}~\cite{Oncescu21a} & 36.1 & 72.0 & 84.5 & 39.6 & 76.8 & 86.7 & 6.7 & 21.6 & 33.2 & 7.0 & 22.7 & 34.6 \\
ML-ACT\textsuperscript{*}~\cite{Mei2021ACT} & 33.9 & 69.7 & 82.6 & 39.4 & 72.0 & 83.9 & 14.4 & 36.6 & 49.9 & 16.2 & 37.6 & 50.2 \\
CLAP~\cite{clap} & 32.7 & 68.0 & 81.2 &	43.9 & 77.7 & 87.6 & 15.6 & 38.6 & 52.3 & 23.7 & 48.9 & 59.9 \\ \hline
\rowcolor{gray!20}\ourmodel & \textbf{37.3} & \textbf{72.8} & \textbf{84.5} & \textbf{49.3} & \textbf{81.8} & \textbf{91.5} & \textbf{17.2} & \textbf{41.0} & \textbf{53.5} & \textbf{23.3} & \textbf{49.5} & \textbf{63.6} \\
\end{tabular}
\vspace{5pt}
\caption{\textbf{Text-to-Audio retrieval} and \textbf{Audio-to-Text retrieval} (R@1,5,10$\uparrow$\label{tab:clap}) on AudioCaps and Clotho. \textsuperscript{*}: models trained without LAION-630K~\cite{clap}.}
\end{table}

As shown above, \ourmodel significantly outperforms CLAP and other recent audio-text models, achieving new state-of-the-art performance on both audio-to-text and text-to-audio retrieval tasks. 
These results further validate the effectiveness of \ourmodel's representations not only in audio-video and audio-only tasks, but also in audio-text tasks.

\section{Limitations and Impacts}\label{sec:app:limit}
\paragraph{Limitations.}
There are several limitations associated with \ourmodel.
Firstly, the scale of the data poses a limitation. 
The AudioSet~\cite{gemmeke2017audio} dataset used by \ourmodel, with two million samples, is approximately two orders of magnitude smaller than the text corpora used in recent language models~\cite{bert,roberta,gpt}.
It is also an order smaller than image corpora like ImageNet-21K used by MBT~\cite{nagrani2021attention}.

Another limitation pertains to the duration of each audio sample. The 10-second recording in AudioSet are relatively short, which can hinder the proper learning of distant temporal dependencies in audio and video. This limitation restricts the potential applicability of \ourmodel to tasks that require modeling longer audio sequences, such as automatic speech recognition (ASR).
Regarding video modeling, due to GPU memory constraints and choice of video footprints, \ourmodel only models 4-second video segments. This limitation makes it challenging to effectively model long video sequences. Additionally, the presence of low-quality videos and misaligned audio-video pairs in AudioSet may adversely affects pre-training.

\paragraph{Potential Societal Impacts.}
The datasets used in this paper, including AudioSet and other end task datasets, were properly licensed and publicly available at the time of data collection. It is important to note that some of the data may have been removed by YouTube or the dataset uploaders. Most of the data in these datasets are licensed under the Creative Commons BY-NC-ND 4.0 license or the Creative Commons 3.0 International License.

To investigate the bias in AudioSet, we selected 200 videos containing speech. In these videos, we did not observe any visual bias in the sampled speakers, which encompassed a wide range of ages, races, and genders. However, it is possible that there may be biases in the distribution of population and ethnicity within AudioSet. It is important to exercise caution and be aware of the potential unintended gender, racial, and societal biases present in AudioSet, which serves as the pre-training data for \ourmodel.

Given that AudioSet consists of a vast collection YouTube videos, there is a potential risk that \ourmodel could learn to reconstruct sensitive personal information, which could then be exploited for malicious purposes, including the creation of audio deepfakes~\cite{SahidullahKH15,chintha2020recurrent}.
To address this concern, the released \ourmodel would be discriminative models, specifically the audio and video encoders, rather than generative models such as decoders. This shift aims to mitigate the potential risks associated with generating synthetic content that could be misused.

\end{document}